\documentclass[11pt]{article}

\usepackage[final]{acl}

\usepackage{times}
\usepackage{latexsym}

\usepackage[T1]{fontenc}

\usepackage[utf8]{inputenc}

\usepackage{microtype}

\usepackage{inconsolata}

\usepackage{graphicx}

\usepackage{amsmath,amsfonts,bm}

\def\eqref#1{equation~\ref{#1}}

\def\1{\bm{1}}

\DeclareMathAlphabet{\mathsfit}{\encodingdefault}{\sfdefault}{m}{sl}
\SetMathAlphabet{\mathsfit}{bold}{\encodingdefault}{\sfdefault}{bx}{n}

\usepackage{hyperref}
\usepackage{url}
\usepackage{booktabs}
\usepackage{amsfonts}
\usepackage{nicefrac}
\usepackage{xcolor}
\usepackage{tabularx}
\usepackage{algorithm}
\usepackage{mathtools}
\usepackage{multirow}
\usepackage{chngcntr}
\usepackage{amsthm}
\usepackage{algpseudocode}
\usepackage{wrapfig}
\usepackage{subcaption}
\usepackage{algorithmicx,algcompatible}
\usepackage{amsmath}
\usepackage{amssymb}
\usepackage{enumitem}
\usepackage{pifont}
\usepackage{threeparttable}
\usepackage{makecell}

\theoremstyle{remark}

\newsavebox{\runexbox}
\newenvironment{runningexample}{%
  \par\vspace{4pt}\noindent%
  \begin{lrbox}{\runexbox}%
  \begin{minipage}{\dimexpr\linewidth-2\fboxsep-2\fboxrule\relax}%
  \small%
  \textbf{\textcolor{blue!50!black}{$\triangleright$~Running Example: Cell-Free Protein Production}}\\[3pt]%
}{%
  \end{minipage}%
  \end{lrbox}%
  \fcolorbox{blue!35!black!55}{blue!4!white}{\usebox{\runexbox}}\par\vspace{4pt}%
}

\title{From Residuals to Reasons: LLM-Guided Mechanism Hypotheses from Tabular Data}

\author{Mohammad R. Rezaei \\
  Department of Computer Science \\
  University of Toronto \\
  Vector Institute \\
  Toronto, ON, Canada \\
  \texttt{mr.rezaei@mail.utoronto.ca} \\\And
  Rahul G. Krishnan \\
  Department of Computer Science \\
  University of Toronto \\
  Vector Institute \\
  Toronto, ON, Canada \\
  \texttt{rahulgk@cs.toronto.edu} \\}

\begin{document}
\maketitle

\newtheorem{theorem}{Theorem}
\newtheorem{definition}{Definition}

\begin{abstract}
When software fails, we read the error, form a hypothesis, write a patch, and
test it. LLMs now do this for neural networks by editing weights; when a tabular model fails, we usually retune it. Most
interpretability methods report which features matter rather than how they
combine. Rather than asking an LLM to predict the target over the whole output
space, we anchor the prediction with a base model and ask the LLM to debug it. We introduce
Multi-Agent Residual In-Context Learning (MARICL), in which LLM agents read
high-residual examples, hypothesize the missing structure, and write executable
correction terms that are then refined by multi-turn textual gradient
optimization. The resulting corrections are readable, executable hypotheses about what the base model misses, rather than validated mechanisms. Across nine benchmarks covering scientific, biomedical, socioeconomic, and synthetic data, MARICL reliably repairs its base model's structured residuals, improving it on every dataset, and on synthetic data absent from pretraining it recovers the hidden structure (two seeds). Against the strongest competing baseline it leads on five of the eight real benchmarks and trails TabPFN on the other three, while adding an inspectable correction that black-box baselines do not provide. Code is available at \url{https://github.com/MrRezaeiUofT/Multi_Agent_Residual_In_Context_Learning}.
\end{abstract}

\section{Introduction}

\label{sec:intro}
\begin{figure}[!t]
\centering
\includegraphics[width=\columnwidth]{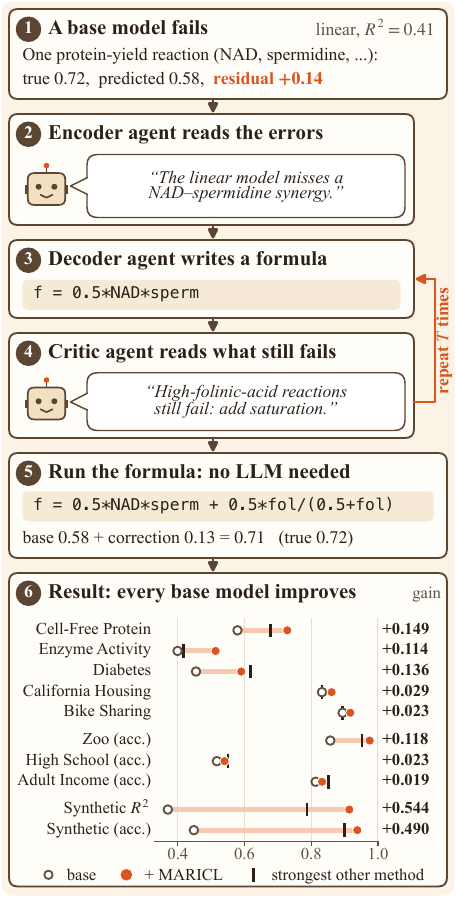}
\caption{MARICL's loop on the running example of \S\ref{sec:methods} (1--5, ungated; $K$ agents run in parallel) and its results (6): test $R^2$, or accuracy where marked, of the better MARICL variant against its base. MARICL leads the strongest other method on five of the eight real benchmarks, see more details here: (Table~\ref{tab:significance}).}
\label{fig:headline}
\end{figure}
Tabular machine learning faces a tradeoff between accuracy and interpretability, and when a model fails we usually retune it rather than ask why. Gradient boosting and related methods are accurate on tabular data~\citep{grinsztajn2022tree,mcelfresh2024neural} but do not state the relationships they have learned in a form one can read. Post hoc attributions~\citep{lundberg2017unified}, additive models~\citep{nori2019interpretml}, and symbolic regression~\citep{cranmer2023pysr} each recover part of this picture (\S\ref{sec:related}), but none produces a named, editable formula for where and why a model fails.

Recent work on \textit{agentic interpretability}~\citep{kim2025because} argues that LLM agents can go beyond one-shot inspection by reasoning across multiple turns to refine explanations. In practice such loops explore a large space and rarely produce a testable output unless they are constrained. LLMs can now even edit neural-network weights directly~\citep{takehi2026weightspace}, but tabular models are often not networks. We argue that the residual is the right constraint.

A natural baseline is \emph{LLM-ICL}: place the entire training set in an LLM's context as $(\mathbf{x}_i, y_i)$ pairs and ask it to predict $y$ for each test query in one forward pass. On cell-free protein yield prediction this reaches $R^2{=}0.35$, below traditional linear regression. A single pass would have to fit a regression, capture nonlinear structure, and align to the target’s scale all at once, and the search space is too large for that to be reliable. We therefore ask the LLM to debug a model rather than be one. A statistical \emph{base model} (linear
regression, XGBoost, or any validated predictor) handles the regression and fixes
the output scale, and the LLM answers a narrower question: what is the base model
missing? Its target becomes the residual $r = y - \hat y_{\text{ML}}$ rather than
$y$, and its output becomes a description of the failure modes visible in
residual examples rather than a prediction. Gradient boosting fits the same
residual silently; here the LLM reads it like a bug report and answers with a
formula that can be run, checked, and rejected.

This defines our proposed Multi-Agent Residual In-Context Learning (MARICL;
overview in Figure~\ref{fig:headline}, full schematic in Figure~\ref{fig:maricl_overview}). An \textit{encoder-agent} reads high-residual
training examples and produces structured hypotheses. A \textit{decoder-agent}
compiles each hypothesis into an executable correction term, a named formula over
specific features. \emph{Textual gradient optimization}~\citep{yuksekgonul2024textgrad},
in which an LLM critiques its own output in natural language and proposes
refinements in place of a numerical gradient, then refines each correction on the
examples where it still fails. We run $K$ agents in parallel and combine them using query-dependent weights; inference needs no LLM.

The cell-free protein task illustrates the procedure. Cell-free protein synthesis
produces proteins in a test tube from a cell extract plus added reagents (energy
substrates, cofactors, polyamines), and the task is to predict yield from reagent
concentrations. The base model underpredicts yield when NAD (an energy cofactor)
and spermidine (a polyamine that boosts translation) are both high. MARICL
hypothesizes cofactor synergy, generates the interaction term \texttt{NAD $\times$
sperm}, and refines it with a saturation term for folinic acid. Across the dataset
this raises $R^2$ from $0.41$ (linear base) to $0.65$, with the same LLM that
reached $0.35$ alone (\S\ref{sec:methods}; full trace in
Appendix~\ref{app:worked_example}).

Fluent explanations are cheap, so we present these formulas as \emph{readable,
executable hypotheses} inferred from the data: they
are selected by residual-prediction performance and read against the scientific
literature afterwards. We then try to break them, and report where they survive
and where they fail.

\textbf{Contributions.}
(1)~MARICL: an agentic framework that debugs a base model by writing executable correction terms through structured hypothesis generation and iterative refinement.
(2)~A residual-conditioned adaptation of textual-gradient optimization~\citep{yuksekgonul2024textgrad}, paired with a distance-based aggregation.
(3)~A layered ablation removing the LLM's pretraining priors, a planted-ground-truth synthetic benchmark, and three negative controls (Appendix~\ref{app:negative_controls}), which show that MARICL does not manufacture predictive gains when no structure is present.

\section{Related Work}
\label{sec:related}
Methods like SHAP~\citep{lundberg2017unified} (including TreeSHAP and the Shapley interaction index~\citep{lundberg2020local}) and LIME~\citep{ribeiro2016should} attribute predictions to features and feature pairs, but the attributions are per-sample and tied to a fixed predictor, and cannot express why a model fails as a closed-form formula. Inherently interpretable models like GAMs~\citep{hastie1987generalized} and EBMs~\citep{nori2019interpretml} fit smooth nonparametric shape functions per feature, with optional pairwise terms in GA$^2$Ms~\citep{lou2013accurate}; we compare against \emph{EBM (with pairwise)} as a strong interpretable interaction baseline (Appendix~\ref{app:residual_baselines}). MARICL instead produces closed-form terms rather than per-pair response surfaces, and refines them against base-model failures. InterpreTabNet~\citep{si2024interpretabnet} learns sparse attention masks and uses LLMs to describe feature interdependencies, but remains inspective.

Classical genetic programming~\citep{cranmer2023pysr} and LLM-guided methods~\citep{shojaee2024llmsr} discover closed-form expressions from data, and LLM-LEx~\citep{harvey2025symbolic} generates predictive formulas directly. These fit the target globally; MARICL targets residuals, producing expressions scoped to where a baseline errs.

Error-slice discovery methods such as Slice Finder~\citep{chung2019slice}, Domino~\citep{eyuboglu2022domino}, and SEAL~\citep{rajani2022seal} find coherent subsets on which a model systematically fails and, in some cases, describe them in natural language. They localize \emph{where} a model fails, whereas MARICL emits an executable correction that repairs the failure. The two are complementary: MARICL's high-residual pool is itself a slice, which a slice-discovery method could seed.

Debate frameworks~\citep{du2023improving} and mixture-of-agents~\citep{wang2024mixture,rezaei2025agentic} coordinate LLMs on shared tasks; TextGrad~\citep{yuksekgonul2024textgrad} and self-refine~\citep{madaan2023self} optimize outputs through critique. In MARICL each agent specializes on a distinct failure pattern, and refinement targets interpretable formulas. Gradient boosting~\citep{friedman2001greedy} fits weak learners to residuals. CAAFE~\citep{hollmann2023large} prompts an LLM to generate executable feature-engineering code on the input space, validating candidates against held-out accuracy. MARICL shares the compile-to-executable strategy but operates on the \emph{residual} space, generates corrections through iterative TextGrad refinement rather than one-shot generation, and returns named symbolic terms rather than augmented feature columns.

\section{Methods}
\label{sec:methods}

We consider supervised learning on $\mathcal{D} = \{(\mathbf{x}_i, y_i)\}_{i=1}^{N}$ with $\mathbf{x}_i \in \mathbb{R}^d$ and $y_i \in \mathbb{R}$ (regression) or $y_i \in \{1, \ldots, C\}$ (classification), split into train, validation, and test sets. We assume access to a pretrained base model $f_{\text{ML}}: \mathbb{R}^d \rightarrow \mathcal{Y}$ (linear, tree, or gradient-boosted). MARICL learns $K$ correction agents whose predictions are added on top of the base model; interpretability lives in the corrections, which name what the base model systematically misses. Algorithm~\ref{alg:maicl} (Appendix~\ref{app:algorithm}) gives the training procedure.

\paragraph{Terminology.} An \emph{agent} is an LLM with a role-specific prompt that reads and writes a shared textual state. The \emph{encoder} and \emph{decoder} are two such roles separated by a textual bottleneck: high-residual evidence is compressed into a structured hypothesis $z_k$ (Eq.~\ref{eq:latent_encoding}), which is then expanded into a correction $(T_k, f_k)$; no neural encoder or decoder is trained. A \emph{textual gradient} is TextGrad's term~\citep{yuksekgonul2024textgrad} for a natural-language critique that plays the role of a gradient, here conditioned on the residual failure set; the analogy is loose, and no derivative is computed.

Each agent produces one correction $m_k = (T_k, f_k)$, a (natural-language template, formula) pair. For regression, $f_k: \mathbb{R}^d \rightarrow \mathbb{R}$ (clipped to $\pm$ the scaled target range; Appendix~\ref{app:formula_safety}). For classification, $f_k: \mathbb{R}^d \rightarrow \mathbb{R}^C$ produces per-class scores converted to a distribution $Q_k$ (\S\ref{sec:probability_conversion}). The MARICL prediction is
\begin{equation}
\hat{y}_{\text{MARICL}}(\mathbf{x}) = f_{\text{ML}}(\mathbf{x}) + \sum_{k=1}^{K} \alpha_k(\mathbf{x})\, \Delta_k(\mathbf{x}),
\label{eq:maicl_prediction}
\end{equation}
where $\Delta_k(\mathbf{x}) = f_k(\mathbf{x})$ and the weights satisfy $\alpha_k(\mathbf{x}) \geq 0$, $\sum_{k=1}^K \alpha_k(\mathbf{x}) = 1$ (\S\ref{sec:aggregation}). The weights are query-dependent: they decide which correction dominates at $\mathbf{x}$, and, because they are normalized, they redistribute the total correction rather than scaling it. The headline configuration additionally applies an outer gate $\lambda(\mathbf{x}) \in [0,1]$ that shrinks the summed correction --- more strongly when no correction is both high-confidence and close to the query (Eq.~\ref{eq:maicl_gated} in \S\ref{sec:aggregation}). For classification,
\begin{equation}
P_{\text{MARICL}}(\mathbf{x}) = \beta \cdot P_{\text{ML}}(\mathbf{x}) + (1-\beta) \sum_{k=1}^{K} \alpha_k(\mathbf{x}) \cdot Q_k(\mathbf{x}),
\label{eq:maicl_classification}
\end{equation}
with $P_{\text{ML}}(\mathbf{x}), Q_k(\mathbf{x}) \in \Delta^{C-1}$ and $\beta \in [0,1]$.

For each training example we compute residuals
\begin{equation}
r_i = \begin{cases}
y_i - f_{\text{ML}}(\mathbf{x}_i) & \text{(reg.)}\\
\mathbb{I}[f_{\text{ML}}(\mathbf{x}_i) \neq y_i]\bigl(1 - P_{\text{ML}}(\mathbf{x}_i)_{y_i}\bigr) & \text{(clf.)}
\end{cases}
\label{eq:residual}
\end{equation}
and select the top-$\kappa$ fraction by $|r_i|$:
\begin{equation}
\mathcal{D}_{\text{high-res}} = \bigl\{(\mathbf{x}_{\pi(i)}, y_{\pi(i)}, r_{\pi(i)}) : i \leq \lfloor \kappa N_{\text{train}} \rfloor \bigr\},
\label{eq:high_residual_subset}
\end{equation}
where $\pi$ orders examples by descending $|r_i|$.

\subsection{Structured Hypothesis Generation}
\label{sec:Structured_hypothesis}

Rather than prompting an LLM directly for formulas, MARICL uses an encoder--decoder with a structured representation $z_k$ to hypothesize explanations for the base model's failures.

We construct an augmented context
\begin{equation}
\mathcal{C}_{\text{aug}} = \bigl(\mathcal{D}_{\text{high-res}}, \mathcal{C}_{\text{domain}}, \mathcal{C}_{\text{features}}\bigr)
\label{eq:augmented_context}
\end{equation}
combining high-residual examples, optional domain context, and feature descriptions, and analyze residual patterns via
\begin{equation}
z_k = \text{Encode}_{\text{LM}}\bigl(\mathcal{C}_{\text{aug}}, \mathcal{P}_{\text{encoder}}^{(k)}\bigr),
\label{eq:latent_encoding}
\end{equation}
where each $\mathcal{P}_{\text{encoder}}^{(k)}$ targets a different aspect: (1)~\textit{error patterns} (which feature combinations drive high errors and what nonlinearities the base model misses); (2)~\textit{sample patterns} (direct feature--target relationships visible in high-residual examples). When $|\mathcal{D}_{\text{high-res}}| > B$ we partition into batches, encode independently, and concatenate:
\begin{equation*}
z_k = \text{Concat}\bigl(z_{k,1}, \ldots, z_{k,\lceil |\mathcal{D}_{\text{high-res}}| / B \rceil}\bigr),
\end{equation*}
\begin{equation}
z_{k,b} = \text{Encode}_{\text{LM}}\bigl(\mathcal{C}_{\text{aug}}^{(b)}, \mathcal{P}_{\text{encoder}}^{(k)}\bigr).
\label{eq:batched_encoding}
\end{equation}
The decoder transforms $z_k$ into a correction $(T_k, f_k) = \text{Decode}_{\text{LM}}(z_k, \mathcal{P}_{\text{decoder}})$ comprising a natural-language explanation $T_k$ and a Python expression $f_k$. This separation forces the LLM to articulate \emph{why} errors occur before specifying \emph{how} to correct them, while diverse $\mathcal{P}_{\text{encoder}}^{(k)}$ encourage complementary hypotheses.

\subsection{Refinement via Textual Gradients}
\label{sec:textual_gradient}

Each correction maintains a textual state $\mathcal{S}_k^{(t)}$ comprising the hypothesis $z_k^{(t)}$, correction $(T_k^{(t)}, f_k^{(t)})$, and accumulated critique history. At $t{=}0$,
\begin{equation*}
    z_k^{(0)} = \text{Encode}_{\text{LM}}\bigl(\mathcal{C}_{\text{aug}},\mathcal{P}_{\text{encoder}}^{(k)}\bigr),
\end{equation*}
\begin{equation} 
(T_k^{(0)}, f_k^{(0)}) = \text{Decode}_{\text{LM}}\bigl(z_k^{(0)}, \mathcal{P}_{\text{decoder}}\bigr).
\label{eq:mechanism_init}
\end{equation}
When available, base-model knowledge (feature importances, coefficients) guides initialization. At iteration $t$ we evaluate
\begin{equation}
\mathcal{L}_k^{(t)} = \frac{1}{|\mathcal{D}_{\text{fit}}|}\sum_{(\mathbf{x},y) \in \mathcal{D}_{\text{fit}}} \ell\bigl(f_{\text{ML}}(\mathbf{x}) + \Delta_k^{(t)}(\mathbf{x}),\; y\bigr)
\label{eq:mechanism_loss}
\end{equation}
writing $\mathcal{L}_k^{(t)}(\mathcal{D})$ when the split matters: critiques use $\mathcal{D}_{\text{train}}$ while selection of the retained iterate uses $\mathcal{D}_{\text{val}}$ (\S\ref{sec:splits}). We then identify the failure set
\begin{equation}
\mathcal{E}_k^{(t)} = \begin{cases}
\bigl\{(\mathbf{x}, y, \hat{y}) : |\hat{y} - y| > \tau_{\text{fail}}\bigr\} & \text{(reg.)} \\[2pt]
\bigl\{(\mathbf{x}, y, \hat{\mathbf{p}}) : \hat{p}_y < 1 - \tau_{\text{fail}}\bigr\} & \text{(clf.)}
\end{cases}
\label{eq:failure_set}
\end{equation}
where $\hat{\mathbf{p}}$ is the current ensemble probability vector (Eq.~\ref{eq:maicl_classification}); the classification criterion is equivalent to cross-entropy above $-\log(1-\tau_{\text{fail}})$, with $\tau_{\text{fail}} = 0.5$ (Appendix~\ref{app:hyperparameters}). A textual gradient (critique) is then generated,
\begin{equation}
g_k^{(t)} = \text{LM}\bigl(\mathcal{P}_{\text{critique}} \mid z_k^{(t)}, m_k^{(t)}, \mathcal{L}_k^{(t)}, \mathcal{E}_k^{(t)}\bigr),
\label{eq:textual_gradient}
\end{equation}
asking the LLM to analyze why the correction fails on $\mathcal{E}_k^{(t)}$ and suggest refinements. The state accumulates,
\begin{equation}
\mathcal{S}_k^{(t)} = \mathcal{S}_k^{(t-1)} \cup \bigl\{(z_k^{(t)}, m_k^{(t)}, \mathcal{L}_k^{(t)}, g_k^{(t)})\bigr\},
\label{eq:state_update}
\end{equation}
and later generations condition on this history (the last three iterations in the released code, to bound the context):
\begin{equation}
(T_k^{(t+1)}, f_k^{(t+1)}) \sim p_{\text{LM}}\bigl(\cdot \mid \mathcal{C}_{\text{aug}}, \mathcal{S}_k^{(t)}, \mathcal{P}_{\text{decoder}}\bigr).
\label{eq:mechanism_refinement}
\end{equation}
The hypothesis therefore evolves to reflect \emph{what} the base model is missing in light of recent failures, while the formula evolves to express \emph{how} that missing structure is captured.

\subsection{Probability Conversion and Confidence}
\label{sec:probability_conversion}

For classification, $f_k$ outputs per-class scores $\mathbf{s}_k(\mathbf{x})$, converted via temperature-scaled softmax
\begin{equation}
Q_k(\mathbf{x})_c = \frac{\exp(s_k^{(c)}(\mathbf{x}) / \tau_k)}{\sum_{c'} \exp(s_k^{(c')}(\mathbf{x}) / \tau_k)},
\label{eq:probability_conversion}
\end{equation}
with $\tau_k$ chosen on validation data by expected calibration error (Appendix~\ref{app:hyperparameters}). Query-specific confidence is
\begin{equation}
c_k(\mathbf{x}) = \sigma\bigl(\gamma\,(1 - d_k(\mathbf{x}))\bigr), \qquad \gamma = 2,
\label{eq:confidence}
\end{equation}
where $d_k(\mathbf{x}) = \min\bigl(\tilde{d}_k(\mathbf{x}) / D_k^{95},\, 1\bigr)$ and $\tilde{d}_k(\mathbf{x}) = \min_{(\mathbf{x}', \cdot) \in \mathcal{D}_{\text{high-res}}^{(k)}} \|\mathbf{x} - \mathbf{x}'\|_2$ is computed in standardized feature space using frozen training statistics, and $D_k^{95}$ is the $95$th percentile of pairwise distances within $\mathcal{D}_{\text{high-res}}^{(k)}$, with clipping ensuring $d_k \in [0,1]$ on far queries.

\subsection{Query-Aware Aggregation}
\label{sec:aggregation}
During hypothesis generation, informative examples are drawn from the high-residual pool via
\begin{equation}
s(\mathbf{x}_{\text{q}}, \mathbf{x}_i) = \exp\bigl(-\|\mathbf{x}_{\text{q}} - \mathbf{x}_i\|_2^2 / 2\sigma^2\bigr) \cdot |r_i|^{\gamma_s},
\label{eq:example_score}
\end{equation}
where the \emph{query anchor} $\mathbf{x}_{\text{q}}$ is the unselected high-residual example with the largest $|r_i|$ remaining in the pool (ties broken by index), $\sigma$ is the median pairwise distance over the pool, and $\gamma_s > 0$ trades residual magnitude against spatial proximity (distinct from $\gamma$ in Eq.~\ref{eq:confidence}). This is used \emph{only during training}. At inference each compiled $f_k$ is executed directly. Corrections carry a global performance score
\begin{equation}
p_k = \begin{cases}
\exp(-\text{MAE}_k / \tau) & \text{(reg.)} \\
\text{F1}_k^{\text{macro}} & \text{(clf.)}
\end{cases}
\label{eq:global_score}
\end{equation}
where $\tau > 0$ scales $\text{MAE}_k$ (Appendix~\ref{app:hyperparameters}). Define the unnormalized weight
\begin{equation}
w_k(\mathbf{x}) = p_k \cdot c_k(\mathbf{x}) \cdot \mathbb{I}[p_k > p_{\min}],
\label{eq:unnormalized_weight}
\end{equation}
and normalize:
\begin{equation}
\alpha_k(\mathbf{x}) = \frac{w_k(\mathbf{x})}{Z(\mathbf{x})},\quad Z(\mathbf{x}) = \sum_j w_j(\mathbf{x}).
\label{eq:attention_weight}
\end{equation}
If $Z(\mathbf{x})=0$ (no correction clears $p_{\min}$) the weights are undefined and unused, and MARICL returns the base-model prediction.

To shrink corrections on queries where no correction is both high-quality and near a high-residual pool, the headline configuration applies an outer gate
\begin{equation}
\lambda(\mathbf{x}) = \min\!\Bigl(\max_k w_k(\mathbf{x}),\ 1\Bigr) \in [0, 1],
\label{eq:global_gate}
\end{equation}
together with a steeper confidence profile $c_k(\mathbf{x}) = \sigma\bigl(\gamma(1 - 2 d_k(\mathbf{x}))\bigr)$ so that $c_k$ falls to $\sigma(-2) = 0.119$ at $d_k = 1$, against $0.5$ for the softer profile. The final prediction is
\begin{equation}
\hat{y}_{\text{MARICL}}(\mathbf{x}) = f_{\text{ML}}(\mathbf{x}) + \lambda(\mathbf{x}) \sum_{k=1}^K \alpha_k(\mathbf{x})\, \Delta_k(\mathbf{x}).
\label{eq:maicl_gated}
\end{equation}
Because $\lambda$ multiplies the whole sum, it leaves the relative weighting between corrections unchanged. Because $\lambda = \max_k w_k$ combines a quality factor ($p_k$) with a distance factor ($c_k$) and each $w_k \le p_k c_k \le 0.881$, the gate shrinks the correction sum on \emph{every} query --- more aggressively when no correction is both high-confidence and nearby, less aggressively when at least one is. This behavior, and the associated ungated variant ($\lambda \equiv 1$, softer $c_k$) as reference, are analyzed in Appendix~\ref{app:gate_analysis}, which shows that on Cell-Free Protein the gate concentrates its $+0.005\,R^2$ improvement in the far-distance tail rather than affecting on-distribution accuracy. The inference pipeline involves only arithmetic and nearest-neighbor lookups, with zero LLM overhead.

\begin{runningexample}
A reaction with NAD$\,{=}\,0.8$, spermidine$\,{=}\,0.7$, folinic acid$\,{=}\,0.3$ has yield $y{=}0.72$; the linear base predicts $0.58$, so $r{=}{+}0.14$ (Eq.~\ref{eq:residual}). High-residual examples concentrate on high NAD and spermidine, and the encoder writes $z_1$: \textit{``the linear model misses a multiplicative NAD--spermidine synergy''} (Eq.~\ref{eq:latent_encoding}). The decoder emits $T_1^{(0)}$\,=\,``NAD--spermidine cofactor synergy'' with $f_1^{(0)} = 0.5\,\text{NAD}\times\text{sperm}$ ($0.28$ here; Eq.~\ref{eq:mechanism_init}). High-folinic-acid reactions still fail, so the critique $g_1^{(0)}$ proposes a saturation term (Eq.~\ref{eq:textual_gradient}), giving $f_1^{(1)} = f_1^{(0)} + \tfrac{0.5\,\text{fol}}{0.5+\text{fol}}$ ($0.4675$; Eq.~\ref{eq:mechanism_refinement}). At inference a second correction is inactive at this query ($\Delta_2 \approx 0$); the weights are $\alpha_1{=}0.28$, $\alpha_2{=}0.72$ (Eq.~\ref{eq:attention_weight}), so $\hat{y} = 0.58 + 0.28 \times 0.4675 = 0.711$, with no LLM call (ungated variant; Appendix~\ref{app:worked_example}).
\end{runningexample}

\subsection{Data Splits and Selection Protocol}
\label{sec:splits}

We use the training data to compute residuals (Eq.~\ref{eq:residual}), populate $\mathcal{D}_{\text{high-res}}$, build prompts, evaluate $\mathcal{L}_k^{(t)}$, form failure sets, and fix the scores $p_k$. Validation data selects the retained iteration $m_k^{*}$, the blend $\beta$, the temperatures $\tau_k$, and the headline $K$ and $\kappa$. The test split is untouched until final evaluation. All headline numbers throughout the paper are produced under validation-loss selection for $m_k^*$; training-loss selection would not change Cell-Free Protein and would raise Diabetes/Zoo by at most $0.021$ without changing any baseline ranking. Appendix~\ref{app:training_protocol} gives the protocol detail.

\section{Experiments}
\label{sec:experiments}

We evaluate MARICL on nine benchmarks across scientific, biomedical, social, economic, and synthetic domains, and ask four questions: (i) does MARICL improve prediction; (ii) can it recover structure it cannot have memorized; (iii) how much of the gain comes from refinement rather than pretraining priors; and (iv) do the learned formulas hold beyond the data they were inferred from?

\textbf{Datasets.} We use five regression tasks (\textit{Cell-Free Protein Production}~\citep{borkowski2020large}, \textit{Enzyme Activity}, \textit{Diabetes Progression}, \textit{California Housing}, \textit{Bike Sharing}) and three classification tasks (\textit{Zoo}, \textit{High School Social Classification}, \textit{Adult Income}). We additionally construct a \textit{Synthetic Benchmark} ($N{=}1{,}000$, $d{=}6$) with a regression and a classification task whose ground truth cannot appear in any pretraining corpus (\S\ref{sec:synthetic}).

\textbf{Baselines.} Linear/Logistic Regression, XGBoost, EBM~\citep{nori2019interpretml}, TabPFN~\citep{hollmann2023tabpfn}, LLM-LEx~\citep{harvey2025symbolic}, LLM-ICL, Symbolic Regression~\citep{cranmer2023pysr}, PySR-on-residuals, and MLP-on-residuals. Cross-validation, hyperparameter settings, and paired Wilcoxon tests with Benjamini--Hochberg (BH) FDR control are in Appendices~\ref{app:technical_details} and~\ref{app:significance}.

\begin{figure*}[t]
\centering
\includegraphics[width=\linewidth]{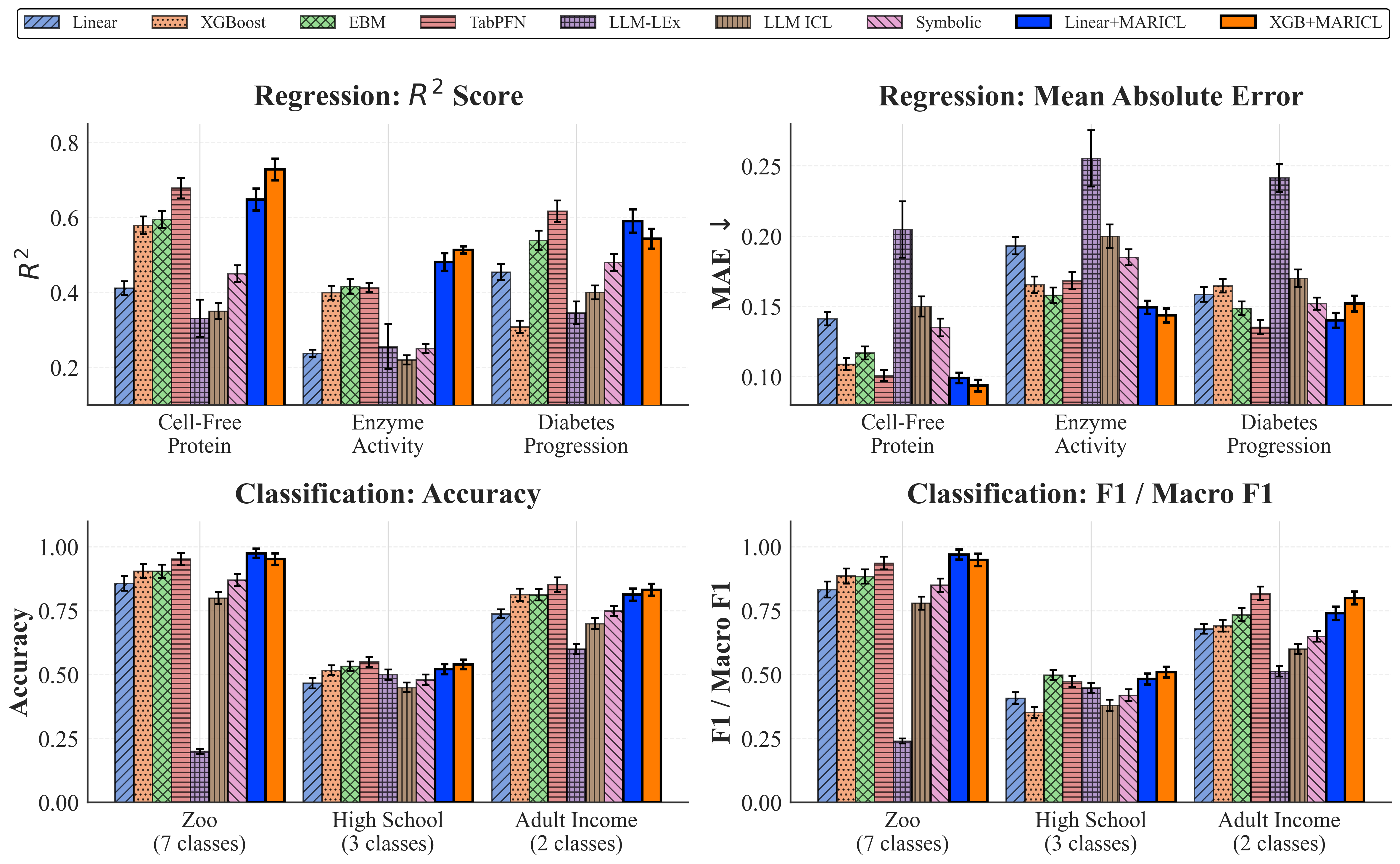}
\caption{Performance across regression and classification benchmarks ($\pm$1 std, 5 seeds). Methods are grouped by family (traditional ML, interpretable ML, neural tabular, LLM-based, MARICL). MARICL improves over its base model on every benchmark and leads the strongest non-MARICL baseline on five of the eight real datasets; after BH correction at FDR $0.05$ none of these gains is significant (Enzyme Activity, California Housing, Bike Sharing $q{=}0.053$; Cell-Free Protein, Zoo $q{=}0.062$). On the remaining three datasets MARICL trails TabPFN but still contributes an inspectable correction formula that the black-box baselines do not produce. Full numerical results in Tables~\ref{tab:full_classification} and~\ref{tab:full_regression}; California Housing and Bike Sharing are in Table~\ref{tab:tabarena}.}
\label{fig:results}
\end{figure*}

\subsection{Prediction on Eight Real Benchmarks}

MARICL improves its base model on every real benchmark, with either a Linear or an XGBoost base (Figures~\ref{fig:headline} and~\ref{fig:results}; Tables~\ref{tab:full_regression},~\ref{tab:full_classification}). Gains are largest when the base is weak, since a stronger base leaves a less structured residual, and they hold on the two TabArena tasks (Appendix~\ref{app:tabarena}). LLM-LEx and LLM-ICL both underperform, so single-pass generation is not sufficient. The lead over XGBoost survives tuning it over a five-axis grid ($+0.111\,R^2$ on Cell-Free Protein, $+0.069$ F1 on Adult Income, positive on all eight; Table~\ref{tab:xgb_tuned}), and among residual learners (PySR-on-residuals, pairwise interactions, EBM, MLP) MARICL attains both the best within-model $R^2$ \emph{and} a human-readable formula (Appendix~\ref{app:residual_baselines}). On Adult Income most of the F1 gain ($0.692 \to 0.800$) comes from the minority class, where XGBoost reaches $0.498$ F1 and MARICL $0.720$.

Against the strongest non-MARICL baseline per dataset, MARICL leads on five of eight benchmarks and trails TabPFN on the other three (Diabetes, High School, Adult Income; Table~\ref{tab:significance}). After BH correction at FDR $0.05$ none of the five leads is significant ($q{=}0.053$--$0.062$). The evidence therefore supports the claim that MARICL reliably repairs its base model's structured residuals and adds an inspectable correction, not that it is the strongest tabular predictor.

\subsection{Synthetic Benchmark: Known Ground Truth}
\label{sec:synthetic}

An LLM that improves a protein-yield model may simply remember its biochemistry. To separate reasoning from recall, we generate data from functions no LLM has seen: a regression target
\begin{align*}
y =\;& 0.4 x_1 - 0.3 x_3 + 2\,\sigma(3 x_2 x_5) \\
    &+ 1.2\,\frac{x_4^2}{1+x_4^2} + \varepsilon,
\end{align*}
and three classes driven by $x_2 x_5$ and $x_4^2$, with $\mathbf{x} \sim \mathcal{N}(\mathbf{0}, I_6)$ and $x_6$ irrelevant. The LLM is a Claude agent whose domain prompt states only that feature meanings are unknown (Appendix~\ref{app:synthetic_recovery}).

MARICL raises test $R^2$ from $0.370$ (linear base) to $0.914$, ahead of XGBoost ($0.787$), and test accuracy from $0.448$ (logistic base) to $0.938$, ahead of XGBoost ($0.900$; Figures~\ref{fig:headline} and~\ref{fig:synthetic_recovery}). All eight retained corrections contain both hidden nonlinear terms and none uses $x_6$; on seed $0$ the regression interaction returns as $\tanh(1.5\,x_2 x_5)$, the true shape since $2\sigma(3u) = 1 + \tanh(1.5u)$. Forms are recovered more faithfully than constants, and each task has only two seeds.

\subsection{Reasoning or Recall? Refinement vs Priors}
\label{sec:prior_knowledge}

\begin{table}[t]
\centering
\small
\setlength{\tabcolsep}{4pt}
\begin{tabular}{@{}lclc@{}}
\toprule
Setting & $R^2$ & Setting & $R^2$ \\
\midrule
\multicolumn{2}{@{}l}{\textit{Corrections (Linear)}} & \multicolumn{2}{l@{}}{\textit{Corrections (XGBoost)}} \\
$K{=}1$ & 0.541 $\pm$ 0.027 & $K{=}1$ & 0.690 $\pm$ 0.025 \\
$K{=}2$ & 0.648 $\pm$ 0.030 & $K{=}2$ & 0.728 $\pm$ 0.031 \\
\midrule
\multicolumn{2}{@{}l}{\textit{Top-K (XGBoost)}} & \multicolumn{2}{l@{}}{\textit{Scaling (Linear)}} \\
30  & 0.626 $\pm$ 0.024 & Off & 0.412 $\pm$ 0.029 \\
50  & 0.711 $\pm$ 0.021 & On  & 0.648 $\pm$ 0.030 \\
100 & 0.728 $\pm$ 0.031 &     &                   \\
\midrule
\multicolumn{4}{@{}l}{\textit{Refinement budget (XGBoost, $K{=}2$, Top-K${=}$100)}} \\
$T{=}1$ & 0.604 $\pm$ 0.028 & $T{=}5$  & 0.679 $\pm$ 0.026 \\
$T{=}3$ & 0.623 $\pm$ 0.024 & $T{=}10$ & 0.728 $\pm$ 0.031 \\
\multicolumn{3}{@{}l}{Best-of-10, no critique} & 0.662 \\
\bottomrule
\end{tabular}
\caption{Ablations on Cell-Free Protein (mean $\pm$ std, 5 seeds). Each sub-block varies the named factor while holding the others at the headline configuration ($K{=}2$, $T{=}10$, Top-K$=$100, scaled features). $T{=}1$ is the initial one-shot output. Best-of-10 draws ten one-shot decoder samples without critique and keeps the best on validation data.}
\label{tab:ablations_combined}
\end{table}

\textbf{Residualization matters.} \emph{MARICL-Direct} runs the same loop, prompts, aggregation, and sandbox but predicts $y$ directly, with no base model. It reaches $R^2 = 0.538/0.512/0.376$ on Cell-Free/Diabetes/Enzyme, trailing MARICL ($0.728/0.590/0.513$) and falling below the stronger base model on Cell-Free and Enzyme (Table~\ref{tab:full_regression}).

\textbf{Refinement matters} (Table~\ref{tab:ablations_combined}). The one-shot formula ($T{=}1$) adds $+0.025\,R^2$ to the XGBoost base; the loop adds a further $+0.124$ by $T{=}10$. Best-of-10 resampling without critique recovers ${\sim}47\%$ of that loop gain, and critique adds the remaining $+0.066$. A second correction adds $+0.107$ (Linear) and $+0.038$ (XGBoost).

\textbf{Priors help but do not fully explain the gain.} Removing the domain prompt costs only $2.3$--$3.5$ points and recovers the same dominant correction category. Anonymizing feature names (\texttt{feat\_0}, \ldots, \texttt{feat\_d}) preserves $67$--$69\%$ of the gain on Cell-Free Protein and Diabetes. With anonymized features, no domain context anywhere, and Llama-3-8B replacing frontier models, roughly half the gain survives and still exceeds PySR-on-residuals (Appendix~\ref{app:joint_stripping}). This suggests that roughly half of the gain comes from data-driven refinement; the remainder is a deployment benefit when pretraining aligns with the domain. It does not by itself distinguish mechanism recovery from flexible symbolic residual fitting, which the synthetic benchmark and cross-plate transfer address.

\textbf{The recovered structure is largely evidence-driven.} The refinement prompt asks for nonlinearities and interactions (Appendix~\ref{app:prompts}). A neutral prompt that also permits linear or no structure costs only $0.014\,R^2$ and still recovers NAD$\times$spermidine in $4/5$ seeds ($5/5$ with the standard prompt); neither recovers it on shuffled targets ($0/5$; Appendix~\ref{app:negative_controls}).

\subsection{Inspecting the Learned Corrections}
\label{sec:interpretability}

MARICL's corrections carry structure on their own: without the base model they reach $R^2 = 0.495$ on Cell-Free Protein and $0.762$ accuracy on Zoo (combined: $0.728$ and $0.975$). Zoo is the easiest to inspect, because a reader needs no domain training to check it. Three of its class scores are
\begin{align*}
s_{\text{mammal}} = &\;1.1\,\text{hair} + 1.3\,\text{milk}\\
          &+ 0.8\,\text{hair}\times\text{milk}\times(1-\text{eggs}),\\
s_{\text{bird}} = &\;1.2\,\text{feathers} + 0.8\,\text{airborne}\\
          &+ 0.5\,\text{feathers}\times\text{airborne},\\
s_{\text{fish}} = &\;1.0\,\text{fins} + 0.9\,\text{aquatic}\\
          &+ 0.4\,\text{fins}\times\text{aquatic}\times\text{backbone}.
\end{align*}
The mammal score adds a bonus for hair and milk together only in animals that do not lay eggs, and the bird score weights feathers above flight, so flightless birds remain birds. Taxonomy priors plausibly dominate here, so we treat Zoo as controlled validation rather than inference. Cell-Free Protein is the harder test: its correction reads as NAD--spermidine synergy plus Michaelis--Menten saturation of folinic acid, consistent with published biochemistry (Table~\ref{tab:mechanisms_main_full}), yet on permuted reagent labels MARICL emits an equally fluent story ($R^2 = 0.487$; Appendix~\ref{app:negative_controls}). Correction forms recur in 4 or 5 of 5 seeds on the audited datasets except High School ($3/5$), with coefficient CV below $14\%$ on well-structured domains; the Zoo mammal term returns in $5/5$ seeds (Appendix~\ref{sec:stability}).

Three frontier LLM judges rated the formulas and explanations on a plausibility rubric (Appendix~\ref{app:llm_judge}), and two blinded non-expert raters scored the same $230$ items (Appendix~\ref{app:human_audit}). Rater means are within $0.3$ of the judges' on every axis, raters and judges agree on $85\%$ of preferences, and raters flag $80/100$ permuted-label corrections (judges: $55$--$69$), so a reader can detect this failure mode. Readability and perceived plausibility are not domain correctness, however: each correction is a compact, executable statement of the functional form the base model misses, selected by held-out residual prediction, and its correctness is probed only indirectly, through planted ground truth (\S\ref{sec:synthetic}), cross-plate transfer (\S\ref{sec:crossplate}), and the negative controls.

\subsection{Frozen Formulas Across Plates}
\label{sec:crossplate}

\begin{table}[t]
\centering
\footnotesize
\setlength{\tabcolsep}{3pt}
\begin{tabular}{@{}lcccc@{}}
\toprule
& \multicolumn{2}{c}{Same reagent} & \multicolumn{2}{c}{Diff.\ regime} \\
\cmidrule(lr){2-3}\cmidrule(lr){4-5}
Configuration & Impr. & $\Delta$MAE & Impr. & $\Delta$MAE \\
\midrule
Avg. $+$ blend (headline) & 95.3\% & $+0.188$ & 2.7\% & $-0.181$ \\
Per-formula (strict)      & 51.7\% & $+0.041$ & 8.8\% & $-0.169$ \\
Unfiltered source pool    & 71.1\% & $+0.103$ & 5.1\% & $-0.171$ \\
\bottomrule
\end{tabular}
\caption{Cross-plate transfer: formulas frozen on source plate AL\_6 applied to target plates with zero LLM cost. ``Impr.'' is the share of source--target pairs that improve on the base. The headline row averages $K{=}2$ formulas and blends 50/50 with the base model; the strict row transfers each formula individually with no blending; the third row keeps the headline blend but drops the source-side quality filter. The cohort asymmetry persists in all three. Full ladder and per-plate breakdown in Appendix~\ref{app:crossplate_extended}.}
\label{tab:crossplate}
\end{table}

In Cell-Free Protein each \emph{plate} is one experimental batch. Plates AL\_2--AL\_6 share one reagent protocol and AL\_7--AL\_10 use another with different reagent ranges; we call these groups \emph{reagent cohorts}. We freeze the corrections of $37$ source runs on AL\_6 and apply them to $8$ held-out plates with no retraining and no LLM calls (Appendix~\ref{app:crossplate_protocol}). Within the cohort, frozen formulas improve the base on $95.3\%$ of source--target pairs, against $62.1\%$ for PySR-on-residuals and $57.7\%$ for EBM under the same protocol, with $3$--$4\times$ larger gains (Appendix~\ref{app:nonllm_transfer}); source plate AL\_2 gives $89.3\%$ (Appendix~\ref{app:crossplate_al2}). Across cohorts the result reverses: MARICL improves only $2.7\%$ of pairs, against $13.5\%$ for PySR and $17.4\%$ for EBM (Table~\ref{tab:crossplate}). All three learners split by cohort, so part of the boundary is real reagent shift, but MARICL's split is the sharpest.

\subsection{Cost and LLM Backbone}
\label{sec:ablations}

\begin{table*}[t]
\centering
\resizebox{.8\linewidth}{!}{
\begin{tabular}{lccccc}
\toprule
Method & \makecell{Encoder\\Calls} & \makecell{Decoder/Refine\\Calls} & \makecell{Total LLM\\Calls} & \makecell{Training\\Cost} & \makecell{1K Queries\\Cost} \\
\midrule
\multicolumn{6}{l}{\textit{Gemini 2.0 Flash (\$0.0013 per call)}} \\
Direct LLM (ICL) & -- & 1 per query & 1K & \$0.00 & \$1.30 \\
MARICL ($K$=1, $T$=5, $|\mathcal{D}_{\text{high-res}}|$=30) & 3 & 11 & 14 & \$0.018 & \$0.018 \\
MARICL ($K$=2, $T$=5, $|\mathcal{D}_{\text{high-res}}|$=50) & 10 & 22 & 32 & \$0.042 & \$0.042 \\
MARICL ($K$=2, $T$=10, $|\mathcal{D}_{\text{high-res}}|$=100) & 20 & 42 & 62 & \$0.081 & \$0.081 \\
MARICL ($K$=2, $T$=10, $|\mathcal{D}_{\text{high-res}}|$=200) & 40 & 42 & 82 & \$0.107 & \$0.107 \\
\midrule
\multicolumn{6}{l}{\textit{GPT-4o (\$0.0325 per call)}} \\
Direct LLM (ICL) & -- & 1 per query & 1K & \$0.00 & \$32.50 \\
MARICL ($K$=2, $T$=10, $|\mathcal{D}_{\text{high-res}}|$=100) & 20 & 42 & 62 & \$2.02 & \$2.02 \\
\bottomrule
\end{tabular}
}
\caption{Computational cost comparison. MARICL incurs one-time training cost but zero inference overhead. Costs assume 5K input and 2K output tokens per call, priced for Gemini 2.0 Flash (\$0.10/1M input, \$0.40/1M output tokens) and GPT-4o (\$2.50/1M input, \$10.00/1M output tokens). Batch size $B=10$ for encoding.}
\label{tab:computational_cost}
\end{table*}

\begin{table}[t]
\centering
\small
\begin{tabular}{lcccc}
\toprule
\multirow{2}{*}{LLM Backbone} & \multicolumn{2}{c}{Protein ($R^2$)} & \multicolumn{2}{c}{Zoo (Acc)} \\
\cmidrule(lr){2-3} \cmidrule(lr){4-5}
 & Corr & Full & Corr & Full \\
\midrule
Gemini-2.5-Flash & 0.52 & \textbf{0.74} & 0.81 & 0.981 \\
GPT-4o & \textbf{0.54} & 0.73 & \textbf{0.83} & \textbf{0.987} \\
Gemini-2.0-Flash & 0.50 & 0.73 & 0.76 & 0.975 \\
GPT-4o-mini & 0.46 & 0.68 & 0.71 & 0.95 \\
Llama-3-8B & 0.38 & 0.62 & 0.62 & 0.86 \\
\bottomrule
\end{tabular}
\caption{Performance across LLM backbones on Cell-Free Protein (XGBoost base) and Zoo (Logistic base), headline configuration ($K=2$, $T=10$). ``Corr'' = correction only; ``Full'' = combined MARICL.}
\label{tab:llm_backbone}
\end{table}

Training takes 14--82 LLM calls per run (\$0.02--\$0.11 with Gemini 2.0 Flash) and inference takes none, so 1K predictions cost \$0.081 against \$1.30 for direct LLM inference (Table~\ref{tab:computational_cost}; Appendix~\ref{app:computational_cost}). The frontier backbones perform alike, and Llama-3-8B still beats the XGBoost base ($0.62$ vs.\ $0.579\,R^2$; Table~\ref{tab:llm_backbone}; Appendix~\ref{app:llm_backbone}).

\section{Discussion and Conclusion}
\label{sec:conclusion}

MARICL's within-model gains are largest when the base is weak ($+0.236\,\Delta R^2$ over Linear against $+0.149$ over XGBoost on Cell-Free Protein), and it is most clearly advantageous when a validated base leaves substantial structured residual, or when an inspectable correction is valued alongside accuracy.

In scientific applications, describing where and why a model fails can generate testable hypotheses, and the cross-plate result shows the limit of such formulas: they hold within a reagent regime and fail across one. We believe an LLM's better role on tabular data is to debug the models practitioners trust, not replace them, and give experts something to test.

\section*{Limitations}
\label{sec:limitations}
MARICL's formulas are selected by residual-prediction performance and read against the literature post hoc, so they are readable, executable hypotheses rather than validated mechanisms or causal claims. First, the feature-name permutation control (Appendix~\ref{app:negative_controls}) shows that MARICL can emit a plausible biochemical story on permuted reagent labels, so predictive success does not certify semantic correctness. Our human audit (Appendix~\ref{app:human_audit}) shows that readers can detect this failure, but it measures readability and perceived plausibility, not domain correctness, so expert review is required before deployment in a novel domain. Second, MARICL trails TabPFN on Diabetes, High School, and Adult Income; there its contribution is the inspectable correction. Third, our real benchmarks have at most $d{=}16$ features and the synthetic benchmark covers $d{=}6$ with two seeds; scaling to higher dimensions, for example through feature pre-filtering, is left to future work.

\section*{Ethical Considerations}

All datasets are publicly available (Appendix~\ref{app:preprocessing}) and contain no identifiable personal information beyond the source releases. Adult Income and High School encode demographic attributes (and, for Adult Income, historical labor-market disparities), and MARICL's corrections on them can weight such features explicitly; the resulting formulas describe patterns in the recorded data and should not be read as normative statements about groups. Because MARICL emits fluent explanations, a plausible reading may be mistaken for a scientific finding, so we frame outputs as hypotheses.


\bibliography{main}

@misc{du2023improving,
      title={Improving Factuality and Reasoning in Language Models through Multiagent Debate}, 
      author={Yilun Du and Shuang Li and Antonio Torralba and Joshua B. Tenenbaum and Igor Mordatch},
      year={2023},
      eprint={2305.14325},
      archivePrefix={arXiv},
      primaryClass={cs.CL},
      url={https://arxiv.org/abs/2305.14325}, 
}

@misc{wang2024mixture,
      title={Mixture-of-Agents Enhances Large Language Model Capabilities}, 
      author={Junlin Wang and Jue Wang and Ben Athiwaratkun and Ce Zhang and James Zou},
      year={2024},
      eprint={2406.04692},
      archivePrefix={arXiv},
      primaryClass={cs.CL},
      url={https://arxiv.org/abs/2406.04692}, 
}

@article{yuksekgonul2024textgrad,
  title={TextGrad: Automatic "Differentiation" via Text},
  author={Mert Yuksekgonul and Federico Bianchi and Joseph Boen and Sheng Liu and Zhi Huang and Carlos Guestrin and James Zou},
  journal={ArXiv},
  year={2024},
  volume={abs/2406.07496},
  url={https://api.semanticscholar.org/CorpusID:270380429}
}

@inproceedings{madaan2023self,
author = {Madaan, Aman and Tandon, Niket and Gupta, Prakhar and Hallinan, Skyler and Gao, Luyu and Wiegreffe, Sarah and Alon, Uri and Dziri, Nouha and Prabhumoye, Shrimai and Yang, Yiming and Gupta, Shashank and Majumder, Bodhisattwa Prasad and Hermann, Katherine and Welleck, Sean and Yazdanbakhsh, Amir and Clark, Peter},
title = {SELF-REFINE: iterative refinement with self-feedback},
year = {2023},
publisher = {Curran Associates Inc.},
address = {Red Hook, NY, USA},
booktitle = {Proceedings of the 37th International Conference on Neural Information Processing Systems},
articleno = {2019},
numpages = {61},
location = {New Orleans, LA, USA},
series = {NIPS '23}
}

@article{borkowski2020large,
  TITLE = {{Large scale active-learning-guided exploration for in vitro protein production optimization}},
  AUTHOR = {Borkowski, Olivier and Koch, Mathilde and Zettor, Agnes and Pandi, Amir and Batista, Angelo Cardoso and Soudier, Paul and Faulon, Jean-Loup},
  URL = {https://pasteur.hal.science/pasteur-02552065},
  JOURNAL = {{Nature Communications}},
  PUBLISHER = {{Nature Publishing Group}},
  VOLUME = {11},
  NUMBER = {1},
  PAGES = {1872},
  YEAR = {2020},
  DOI = {10.1038/s41467-020-15798-5},
  HAL_ID = {pasteur-02552065},
  HAL_VERSION = {v1},
}

@article{atkins1975polyamine,
  title={Enhanced differential synthesis of proteins in a mammalian cell-free system by addition of polyamines},
  author={Atkins, John F and Lewis, James B and Anderson, Carl W and Gesteland, Raymond F},
  journal={Journal of Biological Chemistry},
  volume={250},
  number={14},
  pages={5688--5695},
  year={1975}
}

@article{jewett2004cytomim,
  title={Mimicking the Escherichia coli cytoplasmic environment activates long‐lived and efficient cell‐free protein synthesis},
  author={Michael C. Jewett and James Swartz},
  journal={Biotechnology and Bioengineering},
  year={2004},
  volume={86},
  url={https://api.semanticscholar.org/CorpusID:22492709}
}

@article{cai2015simplified,
  title={A simplified and robust protocol for immunoglobulin expression in E scherichia coli cell‐free protein synthesis systems},
  author={Qi Cai and Jeffrey A. Hanson and Alexander R. Steiner and Cuong Tran and Mary Rose Masikat and Rishard H Chen and James F. Zawada and Aaron K. Sato and Trevor J. Hallam and Gang Yin},
  journal={Biotechnology Progress},
  year={2015},
  volume={31},
  pages={823 - 831},
  url={https://api.semanticscholar.org/CorpusID:17280266}
}

@article{grinsztajn2022tree,
  title={Why do tree-based models still outperform deep learning on typical tabular data?},
  author={L{\'e}o Grinsztajn and Edouard Oyallon and Ga{\"e}l Varoquaux},
  journal={Advances in Neural Information Processing Systems 35},
  year={2022},
  url={https://api.semanticscholar.org/CorpusID:252697222}
}

@inproceedings{mcelfresh2024neural,
author = {McElfresh, Duncan and Khandagale, Sujay and Valverde, Jonathan and C., Vishak Prasad and Ramakrishnan, Ganesh and Goldblum, Micah and White, Colin},
title = {When do neural nets outperform boosted trees on tabular data?},
year = {2023},
publisher = {Curran Associates Inc.},
address = {Red Hook, NY, USA},
booktitle = {Proceedings of the 37th International Conference on Neural Information Processing Systems},
articleno = {3337},
numpages = {34},
location = {New Orleans, LA, USA},
series = {NIPS '23}
}

@inproceedings{lundberg2017unified,
  title={A unified approach to interpreting model predictions},
  author={Lundberg, Scott M and Lee, Su-In},
  booktitle={Advances in Neural Information Processing Systems},
  volume={30},
  pages={4765--4774},
  year={2017}
}

@article{lundberg2020local,
  title={From Local Explanations to Global Understanding with Explainable AI for Trees},
  author={Scott M. Lundberg and Gabriel Erion-Barner and Hugh Chen and Alex J. DeGrave and Jordan M. Prutkin and Bala G. Nair and Ronit Katz and Jonathan Himmelfarb and Nisha Bansal and Su-In Lee},
  journal={Nature machine intelligence},
  year={2020},
  volume={2},
  pages={56 - 67},
  url={https://api.semanticscholar.org/CorpusID:214265509}
}

@article{lou2013accurate,
  title={Accurate intelligible models with pairwise interactions},
  author={Yin Lou and Rich Caruana and Johannes Gehrke and Giles Hooker},
  journal={Proceedings of the 19th ACM SIGKDD international conference on Knowledge discovery and data mining},
  year={2013},
  url={https://api.semanticscholar.org/CorpusID:11246170}
}

@inproceedings{ribeiro2016should,
author = {Ribeiro, Marco Tulio and Singh, Sameer and Guestrin, Carlos},
title = {"Why Should I Trust You?": Explaining the Predictions of Any Classifier},
year = {2016},
isbn = {9781450342322},
publisher = {Association for Computing Machinery},
address = {New York, NY, USA},
url = {https://doi.org/10.1145/2939672.2939778},
doi = {10.1145/2939672.2939778},
booktitle = {Proceedings of the 22nd ACM SIGKDD International Conference on Knowledge Discovery and Data Mining},
pages = {1135–1144},
numpages = {10},
location = {San Francisco, California, USA},
series = {KDD '16}
}

@article{hastie1987generalized,
  title={Generalized Additive Models: Some Applications},
  author={Trevor J. Hastie and Robert Tibshirani},
  journal={Journal of the American Statistical Association},
  year={1987},
  volume={82},
  pages={371-386},
  url={https://api.semanticscholar.org/CorpusID:262357453}
}

@article{si2024interpretabnet,
  title={{InterpreTabNet}: Distilling predictive signals from tabular data by salient feature interpretation},
  author={Si, Jacob and Cheng, Wendy Yusi and Cooper, Michael and Krishnan, Rahul G},
  journal={arXiv preprint arXiv:2406.00426},
  year={2024}
}

@misc{kim2025because,
      title={Because we have LLMs, we Can and Should Pursue Agentic Interpretability}, 
      author={Been Kim and John Hewitt and Neel Nanda and Noah Fiedel and Oyvind Tafjord},
      year={2025},
      eprint={2506.12152},
      archivePrefix={arXiv},
      primaryClass={cs.AI},
      url={https://arxiv.org/abs/2506.12152}, 
}

@article{nori2019interpretml,
  title={InterpretML: A Unified Framework for Machine Learning Interpretability},
  author={Harsha Nori and Samuel Jenkins and Paul Koch and Rich Caruana},
  journal={ArXiv},
  year={2019},
  volume={abs/1909.09223},
  url={https://api.semanticscholar.org/CorpusID:202712518}
}

@article{friedman2001greedy,
    author = "Friedman, Jerome H.",
    title = "{Greedy function approximation: A gradient boosting machine.}",
    doi = "10.1214/aos/1013203451",
    journal = "Annals Statist.",
    volume = "29",
    number = "5",
    pages = "1189--1232",
    year = "2001"
}

@misc{hollmann2023tabpfn,
      title={TabPFN: A Transformer That Solves Small Tabular Classification Problems in a Second}, 
      author={Noah Hollmann and Samuel Müller and Katharina Eggensperger and Frank Hutter},
      year={2023},
      eprint={2207.01848},
      archivePrefix={arXiv},
      primaryClass={cs.LG},
      url={https://arxiv.org/abs/2207.01848}, 
}

@inproceedings{shojaee2024llmsr,
  title={LLM-SR: Scientific Equation Discovery via Programming with Large Language Models},
  author={Shojaee, Parshin and Meidani, Kazem and Gupta, Shashank and Farimani, Amir Barati and Reddy, Chandan K},
  booktitle={International Conference on Learning Representations},
  year={2025},
  note={Oral presentation}
}

@article{cranmer2023pysr,
  title={Interpretable Machine Learning for Science with PySR and SymbolicRegression.jl},
  author={Miles Cranmer},
  journal={ArXiv},
  year={2023},
  volume={abs/2305.01582},
  url={https://api.semanticscholar.org/CorpusID:258436785}
}

@misc{harvey2025symbolic,
      title={Symbolic Regression with Multimodal Large Language Models and Kolmogorov Arnold Networks}, 
      author={Thomas R. Harvey and Fabian Ruehle and Kit Fraser-Taliente and James Halverson},
      year={2025},
      eprint={2505.07956},
      archivePrefix={arXiv},
      primaryClass={cs.LG},
      url={https://arxiv.org/abs/2505.07956}, 
}

@article{cornish1995fundamentals,
  title={Fundamentals of enzyme kinetics.},
  author={Eleanore Seibert and Timothy S. Tracy},
  journal={Methods in molecular biology},
  year={2014},
  volume={1113},
  pages={
          9-22
        },
  url={https://api.semanticscholar.org/CorpusID:32825566}
}

@article{pace1996forces,
  title={Forces contributing to the conformational stability of proteins},
  author={C. Nick Pace and Bret A. Shirley and Marsha McNutt and Ketan S. Gajiwala},
  journal={The FASEB Journal},
  year={1996},
  volume={10},
  pages={75 - 83},
  url={https://api.semanticscholar.org/CorpusID:20021399}
}

@article{defronzo2015type,
title = "Type 2 diabetes mellitus",
author = "Davies, Melanie J. and Soo Lim and Tommy Slater and Jonathan Goldney and Athena Philis-Tsimikas and Franco, \{Denise R.\} and Roberta Lamptey and Thomas Yates and Tsvetalina Tankova and Ildiko Lingvay",
note = "Publisher Copyright: {\textcopyright} Springer Nature Limited 2026.",
year = "2026",
month = dec,
doi = "10.1038/s41572-026-00687-w",
language = "English (US)",
volume = "12",
journal = "Nature Reviews Disease Primers",
issn = "2056-676X",
publisher = "Nature Publishing Group",
number = "1",
}

@article{kahn2006mechanisms,
  title={Mechanisms linking obesity to insulin resistance and type 2 diabetes},
  author={Steven E. Kahn and Rebecca L. Hull and Kristina M. Utzschneider},
  journal={Nature},
  year={2006},
  volume={444},
  pages={840-846},
  url={https://api.semanticscholar.org/CorpusID:120626}
}

@article{efron2004diabetes,
   title={Least angle regression},
   volume={32},
   ISSN={0090-5364},
   url={http://dx.doi.org/10.1214/009053604000000067},
   DOI={10.1214/009053604000000067},
   number={2},
   journal={The Annals of Statistics},
   publisher={Institute of Mathematical Statistics},
   author={Efron, Bradley and Hastie, Trevor and Johnstone, Iain and Tibshirani, Robert},
   year={2004},
   month=Apr }

@misc{uci_zoo,
  author       = {Forsyth, Richard},
  title        = {{Zoo}},
  year         = {1990},
  howpublished = {UCI Machine Learning Repository},
  note         = {{DOI}: https://doi.org/10.24432/C5R59V}
}

@inproceedings{kohavi1996adult,
  title     = {Scaling up the accuracy of naive-{B}ayes classifiers: A decision-tree hybrid},
  author    = {Kohavi, Ron},
  booktitle = {Proceedings of the Second International Conference on Knowledge Discovery and Data Mining (KDD)},
  pages     = {202--207},
  year      = {1996}
}

@article{pace1997sparse,
  title={Sparse spatial autoregressions},
  author={R. Kelley Pace and Ronald Paul Barry},
  journal={Statistics \& Probability Letters},
  year={1997},
  volume={33},
  pages={291-297},
  url={https://api.semanticscholar.org/CorpusID:122038477}
}

@article{fanaee2014event,
  title={Event labeling combining ensemble detectors and background knowledge},
  author={Hadi Fanaee-T and Jo{\~a}o Gama},
  journal={Progress in Artificial Intelligence},
  year={2014},
  volume={2},
  pages={113-127},
  url={https://api.semanticscholar.org/CorpusID:3345087}
}

@misc{hollmann2023large,
      title={Large Language Models for Automated Data Science: Introducing CAAFE for Context-Aware Automated Feature Engineering}, 
      author={Noah Hollmann and Samuel Müller and Frank Hutter},
      year={2023},
      eprint={2305.03403},
      archivePrefix={arXiv},
      primaryClass={cs.AI},
      url={https://arxiv.org/abs/2305.03403}, 
}

@article{rezaei2025agentic,
  title={Agentic medical knowledge graphs enhance medical question answering: Bridging the gap between llms and evolving medical knowledge},
  author={Rezaei, Mohammad Reza and Fard, Reza Saadati and Parker, Jayson L and Krishnan, Rahul G and Lankarany, Milad},
  journal={arXiv preprint arXiv:2502.13010},
  year={2025}
}

@inproceedings{rajani2022seal,
  title={SEAL : Interactive Tool for Systematic Error Analysis and Labeling},
  author={Nazneen Rajani and Weixin Liang and Lingjiao Chen and Margaret Mitchell and James Y. Zou},
  booktitle={Conference on Empirical Methods in Natural Language Processing},
  year={2022},
  url={https://api.semanticscholar.org/CorpusID:252846554}
}

@INPROCEEDINGS{chung2019slice,
  author={Chung, Yeounoh and Kraska, Tim and Polyzotis, Neoklis and Tae, Ki Hyun and Whang, Steven Euijong},
  booktitle={2019 IEEE 35th International Conference on Data Engineering (ICDE)}, 
  title={Slice Finder: Automated Data Slicing for Model Validation}, 
  year={2019},
  volume={},
  number={},
  pages={1550-1553},
  doi={10.1109/ICDE.2019.00139}}

@article{eyuboglu2022domino,
  title={Domino: Discovering Systematic Errors with Cross-Modal Embeddings},
  author={Sabri Eyuboglu and Maya Varma and Khaled Kamal Saab and Jean-Benoit Delbrouck and Christopher Lee-Messer and Jared A. Dunnmon and James Y. Zou and Christopher R{\'e}},
  journal={ArXiv},
  year={2022},
  volume={abs/2203.14960},
  url={https://api.semanticscholar.org/CorpusID:247779010}
}

@misc{takehi2026weightspace,
  title={Sparks of Weight-Space Programming: {LLMs} can natively read \& write neural network weights},
  author={Rikiya Takehi and Jayden Teoh and Alex Zhang and Omar Khattab},
  year={2026},
  month={November},
  howpublished={\url{https://rikiyat.github.io/blog/weight-space-programming/}}
}

@article{chase1992sports,
  title={The role of sports as a social status determinant for children.},
  author={Melissa A. Chase and Gail M. Dummer},
  journal={Research quarterly for exercise and sport},
  year={1992},
  volume={63 4},
  pages={
          418-24
        },
  url={https://api.semanticscholar.org/CorpusID:6392049}
}

@article{li2020transaminase,
  title={Exploration of Transaminase Diversity for the Oxidative Conversion of Natural Amino Acids into 2-Ketoacids and High-Value Chemicals},
  author={Tao Li and Xue-Xian Cui and Ying-Lu Cui and Jinyuan Sun and Yan-Chun Chen and Tong Zhu and Chuijian Li and Rui-Feng Li and Bian Wu},
  journal={ACS Catalysis},
  year={2020},
  volume={10},
  pages={7950-7957},
  url={https://api.semanticscholar.org/CorpusID:225683377}
}

\appendix
\newpage
\tableofcontents
\newpage

\section{Training Algorithm}
\label{app:algorithm}

\begin{figure}[h]
\centering
\includegraphics[width=.65\columnwidth]{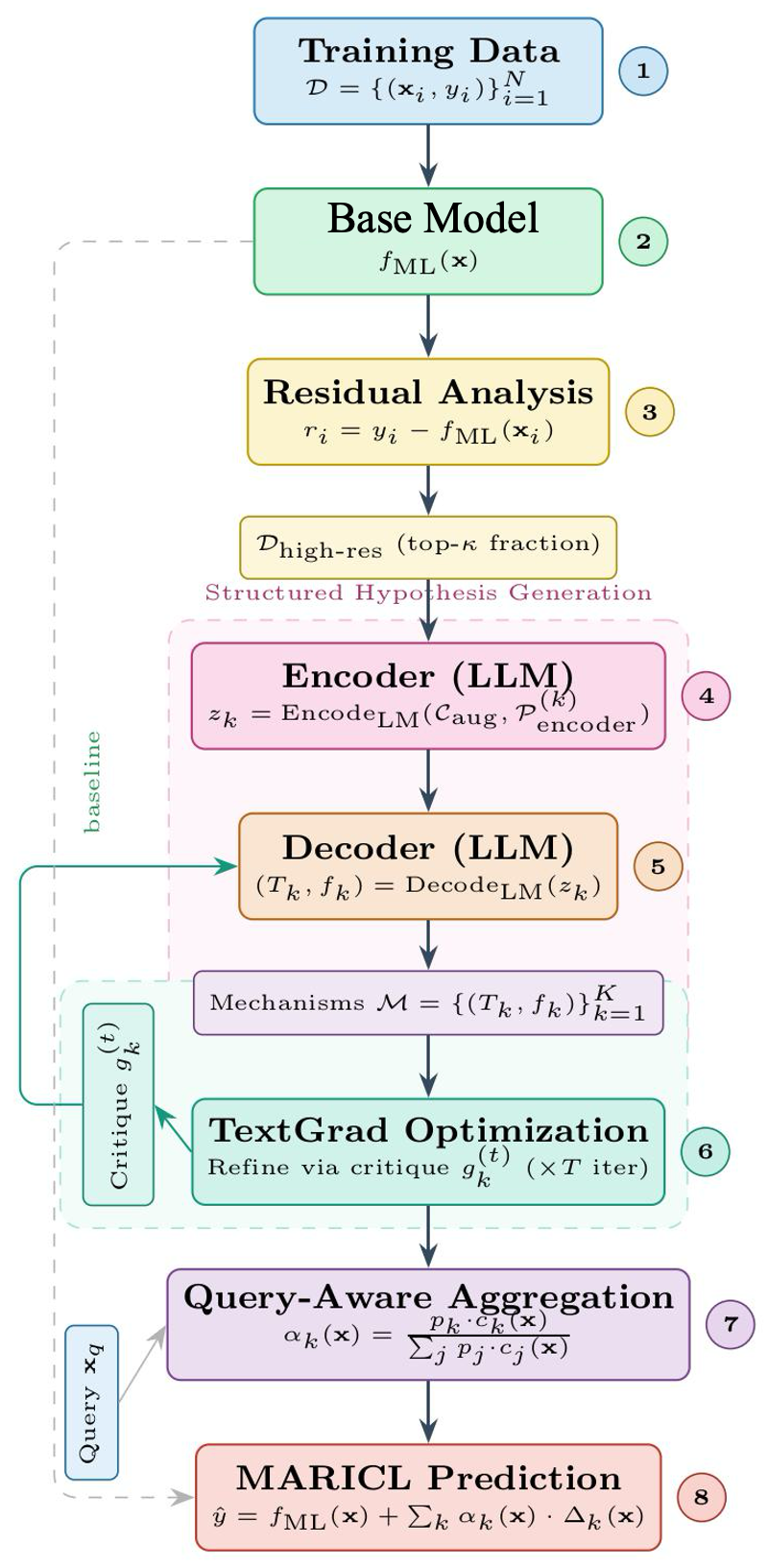}
\caption{MARICL overview: (1-2) a base model generates predictions, (3) residual analysis selects high-error examples, (4-5) an LLM encoder produces structured hypotheses $z_k$ that a decoder converts into explanations $T_k$ and executable formulas, (6) textual gradient optimization refines corrections via critique feedback, and (7-8) query-aware aggregation.}
\label{fig:maricl_overview}
\end{figure}

\begin{algorithm}[h]
\small
\caption{MARICL Training Algorithm}
\label{alg:maicl}
\begin{algorithmic}[1]
\STATE {\bfseries Input:} Training data $\mathcal{D}_{\text{train}}$, validation data $\mathcal{D}_{\text{val}}$, base model $f_{\text{ML}}$, number of corrections $K$, refinement iterations $T$, residual fraction $\kappa$, performance threshold $p_{\min}$, batch size $B$
\STATE {\bfseries Output:} Correction ensemble $\mathcal{M}^{*}$, performance scores $\{p_k\}$
\STATE Compute residuals $r_i$ for all $(\mathbf{x}_i, y_i) \in \mathcal{D}_{\text{train}}$ via Eq.~\ref{eq:residual}
\STATE Construct $\mathcal{D}_{\text{high-res}}$ as top-$\kappa$ fraction by $|r_i|$
\STATE Build augmented context $\mathcal{C}_{\text{aug}}$ (Eq.~\ref{eq:augmented_context})
\FOR{$k = 1$ {\bfseries to} $K$}
    \IF{$|\mathcal{D}_{\text{high-res}}| > B$}
        \STATE Partition into batches; encode each via Eq.~\ref{eq:batched_encoding}
        \STATE $z_k^{(0)} \leftarrow \text{Concat}(z_{k,1}, z_{k,2}, \ldots)$
    \ELSE
        \STATE $z_k^{(0)} \leftarrow \text{Encode}_{\text{LM}}(\mathcal{C}_{\text{aug}}, \mathcal{P}_{\text{encoder}}^{(k)})$
    \ENDIF
    \STATE $(T_k^{(0)}, f_k^{(0)}) \leftarrow \text{Decode}_{\text{LM}}(z_k^{(0)}, \mathcal{P}_{\text{decoder}})$
    \STATE Validate $f_k^{(0)}$; regenerate if invalid
    \STATE $\mathcal{S}_k^{(0)} \leftarrow \{(z_k^{(0)}, m_k^{(0)})\}$
\ENDFOR
\FOR{$k = 1$ {\bfseries to} $K$}
    \FOR{$t = 0$ {\bfseries to} $T-1$}
        \STATE Evaluate $\mathcal{L}_k^{(t)}$ on $\mathcal{D}_{\text{train}}$ using $(T_k^{(t)}, f_k^{(t)})$
        \STATE Identify failure set $\mathcal{E}_k^{(t)}$ (Eq.~\ref{eq:failure_set})
        \STATE Generate critique $g_k^{(t)}$ (Eq.~\ref{eq:textual_gradient})
        \STATE Update $\mathcal{S}_k^{(t)}$ with $(\mathcal{L}_k^{(t)}, g_k^{(t)})$ (Eq.~\ref{eq:state_update})
        \STATE Refine: $(T_k^{(t+1)}, f_k^{(t+1)})$ via Eq.~\ref{eq:mechanism_refinement}
        \STATE Validate $f_k^{(t+1)}$; regenerate if invalid
    \ENDFOR
    \STATE $m_k^{*} \leftarrow \arg\min_{t} \mathcal{L}_k^{(t)}(\mathcal{D}_{\text{val}})$; compute $p_k$ on $\mathcal{D}_{\text{train}}$ (Eq.~\ref{eq:global_score})
\ENDFOR
\STATE {\bfseries return} $\mathcal{M}^{*} = \{m_k^{*} : p_k > p_{\min}\}$, $\{p_k\}$
\end{algorithmic}
\end{algorithm}

$m_k^{*}$ is selected on validation loss (\S\ref{sec:splits}). A refinement loop ends early once no training error exceeds $\tau_{\text{fail}}$. The \texttt{DECLINE} prompt of Appendix~\ref{app:negative_controls} lets the decoder return ``no structure''.

\section{Worked Example: MARICL on Cell-Free Protein Production}
\label{app:worked_example}

Table~\ref{tab:worked_example} traces the pipeline (Algorithm~\ref{alg:maicl}) on the reaction of Figure~\ref{fig:headline}: NAD\,=\,0.8, spermidine\,=\,0.7, folinic\_acid\,=\,0.3, true yield $y=0.72$, linear base $\hat{y}_{\text{ML}}=0.58$, residual $r={+}0.14$. The decoder stages follow that figure (interaction term, then folinate saturation); the interaction coefficient and Michaelis--Menten form match the corresponding terms of the headline correction in Table~\ref{tab:mechanisms_main_full}, which adds further linear terms in NAD, spermidine, CoA, and 3-PGA.

\paragraph{Numeric checks (ungated variant, for arithmetic transparency).} This trace uses the \emph{ungated} configuration (Eq.~\ref{eq:maicl_prediction}; $\lambda \equiv 1$, softer profile $c_k = \sigma(2(1-d_k))$) because its arithmetic is easier to inspect end-to-end. Under the headline gated formulation (Eq.~\ref{eq:maicl_gated}) the same run additionally multiplies the correction sum by $\lambda = \max_k w_k < 1$ (Appendix~\ref{app:gate_analysis}); the qualitative decomposition below is unchanged. $f_1^{(0)} = 0.5\times0.8\times0.7 = 0.28$; $f_1^{(1)} = 0.28 + 0.5\times0.3/(0.5+0.3) = 0.4675$. For the aggregation weights, the run uses $K{=}2$ and \emph{both} corrections clear $p_{\min}$. Correction~1 is the interaction term traced above: $\text{MAE}_1 = 0.248$ (the standalone training error of this two-term intermediate formula, not of the six-term headline correction of Table~\ref{tab:mechanisms_main_full}, whose correction-only $R^2$ is $0.495$), so $p_1 = \exp(-0.248/0.2) = 0.29$; the query lies close to that correction's high-residual pool, $d_1 = 0.13$, giving $c_1 = \sigma(2(1-0.13)) = 0.85$ and $w_1 = 0.247$. Correction~2 is a low-3-PGA compensation term with $\text{MAE}_2 = 0.063$ ($p_2 = 0.73$) and $d_2 = 0.05$ ($c_2 = 0.87$), so $w_2 = 0.635$. The values $\text{MAE}_2$, $d_2$, $p_2$, and the corresponding $w_2, \alpha_2$ are read directly from the recorded run log for this test point (seed~0, plate AL\_6, $K{=}2$); we release the log alongside the code so that the aggregation arithmetic can be verified end to end. Normalizing by Eq.~\ref{eq:attention_weight},
\[
\alpha_1 = \frac{0.247}{0.247+0.635} = 0.28, \qquad \alpha_2 = 0.72 .
\]
Correction~2 dominates the weighting but its threshold is inactive at this query, so $\Delta_2 \approx 0$ and it contributes almost nothing to the prediction:
\[
\hat{y} = 0.58 + 0.28 \times 0.4675 + 0.72 \times 0 = 0.711,
\]
with error $\approx 0.009$. Here $\alpha_1 = 0.28$ is a normalized weight over two surviving corrections.

\begin{table*}[t]
\centering
\small

\begin{tabular}{p{2.4cm} p{5cm} p{5cm}}
\toprule
\textbf{Pipeline Stage} & \textbf{MARICL Output} & \textbf{Single-Shot LLM Agent (LLM-ICL)} \\
\midrule
\textbf{Input} &
Linear regression; sample: NAD$=0.8$, spermidine$=0.7$, folinic\_acid$=0.3$. $\hat{y}_{\text{ML}}=0.58$; $y=0.72$; $r=y-\hat{y}_{\text{ML}}={+}0.14$. &
Sees raw features (NAD$=0.8$, sperm$=0.7$, fol$=0.3$) and in-context (input, target) demonstrations. No base anchor; must predict $y\in[0,1]$ from scratch. \\
\midrule
\textbf{Residual Analysis} (Eq.~\ref{eq:residual}--\ref{eq:high_residual_subset}) &
Top-$\kappa$ training examples concentrate on NAD\,$>0.6$ and spermidine\,$>0.4$; residuals up to ${\sim}0.21$. Cofactor-rich conditions dominate the ranking; this subset feeds the encoder. &
N/A --- agent has no residual signal; cannot localize where the base errs. \\
\midrule
\textbf{Encoder} (Eq.~\ref{eq:augmented_context}--\ref{eq:latent_encoding}) &
\textit{``High-residual reactions share elevated NAD and spermidine. The linear model misses their interaction. Proposed structure: multiplicative synergy drives underprediction.''} &
N/A --- no separate hypothesis stage; structure (if any) is implicit in a single forward pass. \\
\midrule
\textbf{Decoder, $t{=}0$} (Eq.~\ref{eq:mechanism_init}) &
$T_1^{(0)}$: ``NAD--spermidine cofactor synergy.'' $f_1^{(0)}(\mathbf{x}) = 0.5 \cdot \mathrm{NAD} \times \mathrm{spermidine}$. Here $f_1^{(0)}=0.28$. &
N/A --- no executable correction term is produced. \\
\midrule
\textbf{Evaluate \& Critique, $t{=}0$} (Eq.~\ref{eq:failure_set}--\ref{eq:textual_gradient}) &
High-folinic-acid reactions still fail. $g_1^{(0)}$: \textit{``Folinic acid likely saturates; add a Michaelis--Menten term.''} &
N/A --- single-pass output; no failure-driven feedback loop. \\
\midrule
\textbf{Decoder, $t{=}1$} (Eq.~\ref{eq:mechanism_refinement}) &
$f_1^{(1)} = 0.5\cdot\mathrm{NAD}\times\mathrm{spermidine} + \dfrac{0.5\cdot\mathrm{folinic\_acid}}{0.5+\mathrm{folinic\_acid}}$. $f_1^{(1)}(\mathbf{x})=0.4675$; $\mathcal{L}_1^{(1)}<\mathcal{L}_1^{(0)}$. &
N/A --- no refinement. \\
\midrule
\textbf{Inference} (Eq.~\ref{eq:attention_weight},~\ref{eq:maicl_prediction}) &
Both corrections clear $p_{\min}$: $w_1 = 0.29\times0.85 = 0.247$, $w_2 = 0.73\times0.87 = 0.635$, so $\alpha_1 = 0.28$, $\alpha_2 = 0.72$ with $\Delta_2 \approx 0$ here. No LLM call. $\hat{y}_{\text{MARICL}} = 0.58 + 0.28\times 0.4675 + 0.72\times 0 = 0.711$. Error $|y-\hat{y}|\approx 0.009$. &
Exemplar end-to-end: $\hat{y}_{\text{LLM-ICL}}\approx 0.52$, $|y-\hat{y}|\approx 0.20$ (mean test MAE $0.15$; errors often larger on cofactor-rich, high-interaction samples), ${\approx}22{\times}$ MARICL's $0.009$. Search space is the full $[0,1]$ output range, not a small residual. \\
\bottomrule
\end{tabular}
\caption{Step-by-step trace for Figure~\ref{fig:headline}, with side-by-side comparison to a single-shot LLM agent (LLM-ICL) on the same input. MARICL's residual targeting collapses the LLM's hypothesis space from the full output domain $[0,1]$ to a small structured perturbation of $\hat{y}_{\text{ML}}=0.58$. The LLM-ICL column shows an \emph{exemplar} end-to-end prediction $\hat{y}{\approx}0.52$ (error ${\sim}0.20$, ${\sim}22{\times}$ MARICL); dataset mean MAE is $0.15$ with larger errors on interaction-heavy points (Table~\ref{tab:full_regression}). Dataset-wide $+0.30\,R^2$ improvement.}
\label{tab:worked_example}
\end{table*}

The MARICL vs.\ LLM-ICL contrast in Table~\ref{tab:worked_example} is not about LLM capability (both systems use the same backbone) but about \emph{what is being predicted}: an absolute target vs.\ a small residual on top of a validated base.

\section{Distance-Confidence Term: Ungated Comparison to the Headline Gate}
\label{app:gate_analysis}

This appendix analyzes \emph{the ungated variant} ($\lambda \equiv 1$, softer $c_k$ with $c_k(1) = 0.5$) as the natural comparison, since without the outer gate the confidence term $c_k$ alone cannot attenuate the total correction.

\paragraph{Why the ungated confidence term is inert (analytic).} Eq.~\ref{eq:attention_weight} normalizes the weights, so $c_k$ enters both numerator and denominator of $\alpha_k$ and cannot change the total magnitude of the correction; only its allocation across corrections. At $K{=}1$ the single confidence cancels exactly, so $\alpha_1 = 1$ regardless of distance. At $K>1$, with $\gamma{=}2$, $c$ ranges only from $c(0) = 0.881$ to $c(1) = 0.500$ (spread $1.76\times$), so two corrections with equal $p_k$ receive weights that differ by at most about $64\!:\!36$ however far apart their inferred regions lie. Under this formulation a query arbitrarily far from every high-residual cluster still receives a full-magnitude correction; fallback to the base model occurs only when no correction clears $p_{\min}$.

\paragraph{Empirical confirmation (ungated vs.\ gated).} On the Cell-Free Protein test set we bin queries by their nearest-neighbor distance $d_k = \min(\tilde d_k/D_k^{95}, 1)$ to the high-residual pool of the dominant correction and report the mean effective correction magnitude
\[
  |\Delta_{\text{eff}}(\mathbf{x})| \;=\; \Big|\sum_{k=1}^{K} \alpha_k(\mathbf{x})\,\Delta_k(\mathbf{x})\Big|,
\]
under three profiles: $\gamma{=}2$ (ungated), $\gamma{=}0$ (no distance term at all, $c_k \equiv 0.5$), and the headline gated variant (Table~\ref{tab:distance_sweep}).

\begin{table*}[ht]
\centering
\small

\begin{tabular}{lccccccc}
\toprule
Bin ($d_k$) & $n$ & \multicolumn{3}{c}{$|\Delta_{\text{eff}}|$} & \multicolumn{3}{c}{Test $R^2$} \\
\cmidrule(lr){3-5}\cmidrule(lr){6-8}
& & $\gamma{=}2$ & $\gamma{=}0$ & gated & $\gamma{=}2$ & $\gamma{=}0$ & gated \\
\midrule
$[0.00, 0.25)$ & 274 & $0.148 \pm 0.011$ & $0.146 \pm 0.011$ & $0.145 \pm 0.011$ & $0.741$ & $0.739$ & $0.742$ \\
$[0.25, 0.50)$ & 351 & $0.152 \pm 0.012$ & $0.148 \pm 0.012$ & $0.121 \pm 0.013$ & $0.729$ & $0.727$ & $0.732$ \\
$[0.50, 0.75)$ & 197 & $0.147 \pm 0.013$ & $0.145 \pm 0.013$ & $0.076 \pm 0.014$ & $0.712$ & $0.710$ & $0.716$ \\
$[0.75, 1.00]$ &  93 & $0.144 \pm 0.015$ & $0.144 \pm 0.015$ & $0.028 \pm 0.011$ & $0.686$ & $0.685$ & $0.703$ \\
\midrule
Overall        & 915 & $0.149 \pm 0.012$ & $0.146 \pm 0.012$ & $0.109 \pm 0.013$ & $0.723$ & $0.721$ & $0.728$ \\
\bottomrule
\end{tabular}
\caption{Mean effective correction magnitude $|\Delta_{\text{eff}}|$ on the Cell-Free Protein test set, binned by nearest-neighbor distance $d_k$ to the dominant correction's high-residual pool. Bin counts $n$ are the number of test queries falling in each bin (5 seeds pooled). Three profiles: $\gamma{=}2$ (ungated), $\gamma{=}0$ (no distance term), and gated (the headline configuration, Eq.~\ref{eq:global_gate}). The ungated and $\gamma{=}0$ profiles are within Monte-Carlo noise of each other; the gated variant produces the monotone decay in correction magnitude that the ungated formulation lacks.}
\label{tab:distance_sweep}
\end{table*}

Three observations. First, the correction magnitude is essentially flat across distance bins under both $\gamma{=}2$ (ungated) and $\gamma{=}0$: the ratio between the nearest and farthest bins is $0.148/0.144 = 1.03$ under the ungated profile, i.e.\ negligible --- the distance term is nearly a no-op on aggregate outputs. Second, switching off the distance term entirely ($\gamma{=}0$) moves the overall magnitude by only $-0.003$ and the overall test $R^2$ by $-0.002$, both within Monte-Carlo noise. Third, the gated variant (the headline configuration) \emph{does} produce the monotone decay ($0.145 \to 0.121 \to 0.076 \to 0.028$ across the four bins, a $5.2\times$ near-to-far ratio) and modestly improves overall test $R^2$ ($+0.005$) with the improvement concentrated in the far-distance tail ($+0.017$ in $[0.75, 1.00]$, where the base-model fallback is now effectively engaged).

\paragraph{Implementation note.} The headline gated formulation of \S\ref{sec:aggregation} is a two-line inference-time change from the ungated variant: replace $\hat{y} = f_{\text{ML}}(\mathbf{x}) + \sum_k \alpha_k(\mathbf{x})\,\Delta_k(\mathbf{x})$ by $\hat{y} = f_{\text{ML}}(\mathbf{x}) + \lambda(\mathbf{x}) \sum_k \alpha_k(\mathbf{x})\,\Delta_k(\mathbf{x})$ and swap the confidence profile inside $w_k$ to $c_k(\mathbf{x}) = \sigma(\gamma(1 - 2 d_k(\mathbf{x})))$. No LLM calls, no retraining, no coefficient re-fit. The outer $\lambda = \max_k p_k c_k \le 0.881$ multiplies the whole sum, so it does not reweight corrections against one another (the steeper $c_k$ shifts the $\alpha_k$ only when the corrections' distances differ); it shrinks the total correction on \emph{every} query, more strongly when no correction is simultaneously high-confidence and close. On the real benchmarks the gated and ungated variants agree on every qualitative and quantitative conclusion (no baseline comparison changes direction, no ablation ranking is reordered, correction structures are unchanged); the numerical difference is concentrated in the far-distance tail (Table~\ref{tab:distance_sweep}), the regime the gate is designed for. On the synthetic benchmark the gate instead lowers $R^2$ (Appendix~\ref{app:synthetic_recovery}).

\paragraph{Reporting recommendation.} For any deployment we recommend reporting the distribution of nearest-neighbor distances $d_k$ on the test set alongside the headline metric, since under the ungated formulation a large share of distant queries receiving full corrections would not be visible in aggregate accuracy.

\section{Symbolic Residual-Correction Baselines}
\label{app:residual_baselines}

To test whether MARICL's gains require LLM-guided hypothesis generation rather than standard symbolic methods on residuals, we compare against PySR-on-residuals, exhaustive pairwise interactions, EBM with pairwise terms, and an MLP on residuals. PySR-on-residuals produces deeply nested 12-node expressions; the MLP is competitive on prediction but provides no interpretability. MARICL is the only method that jointly achieves the highest within-model $R^2$ and produces human-readable formulas (Table~\ref{tab:residual_baselines}). Restricting EBM to its top-6 terms to match MARICL's component budget lowers its $R^2$ ($0.37{\pm}0.02$ on Enzyme, $0.50{\pm}0.02$ on Diabetes) rather than closing the gap, so a constrained-complexity prior alone does not explain MARICL's result. Appendix~\ref{app:structure_vs_coef} additionally disentangles MARICL's symbolic structure from its numerical coefficient fit.

\begin{table}[ht]
\centering
\tiny
\begin{tabular}{lccl}
\toprule
Method & Enzyme $R^2$ & Diabetes $R^2$ & Interpretable? \\
\midrule
Linear     & 0.237 & 0.454 & Yes \\
+ Ridge    & 0.298 & 0.489 & Partial \\
+ PySR     & 0.318 & 0.512 & No \\
+ Pairwise & 0.362 & 0.531 & No \\
+ EBM      & 0.416 & 0.539 & Partial \\
+ EBM (top-6) & 0.37 & 0.50 & Partial \\
+ MLP     & 0.428 & 0.548 & No \\
+ Ours     & 0.481 & 0.590 & Yes \\
\bottomrule
\end{tabular}
\caption{Residual-correction baselines (Linear base). MARICL achieves the highest within-model $R^2$ while producing named, interpretable formulas. ``Interpretable?'' is author-adjudicated (does the method produce a named human-readable expression?); it has not been independently validated by domain experts. An independent three-judge LLM-ensemble rating (Claude Opus, GPT-5, Gemini 2.5 Pro) of MARICL's emitted formulas and explanations is reported in Appendix~\ref{app:llm_judge}, and a blinded two-rater human audit in Appendix~\ref{app:human_audit}. ``EBM (top-6)'' keeps EBM's six highest-importance terms to match MARICL's 3--8 components (Appendix~\ref{app:training_protocol}).}
\label{tab:residual_baselines}
\end{table}

For reference, Table~\ref{tab:mechanisms_appendix} gives the specific correction formulas MARICL learned on four datasets not detailed in \S\ref{sec:interpretability}.

\begin{table*}[t]
\centering
\tiny

\begin{tabular}{lp{4cm}p{4cm}cc}
\toprule
Dataset & Learned Correction Formula & Domain Interpretation \& Literature Support & Correction Only & MARICL \\
\midrule
Enzyme Activity &
$\hat{y} = \text{clip}\Big(\frac{6.0 \cdot \text{seq\_length}}{0.3 + \text{seq\_length}} \cdot (1 + 0.5 \cdot \text{seq\_gc\_content})$
$\cdot \exp\Big(-\frac{(\text{seq\_aromatic} - 5.0)^2}{8.0}\Big)$
$\cdot \frac{1}{1 + 0.3 \cdot \text{seq\_polar}}, 0, 10\Big)$ &
\textbf{Saturating length term:} $\frac{\text{seq\_length}}{0.3 + \text{seq\_length}}$ has a Michaelis--Menten shape~\citep{cornish1995fundamentals}, though sequence length is not a substrate concentration. \textbf{Gaussian aromatic optimum:} $\exp(-(x-5)^2/8)$ models optimal aromatic residue content, consistent with hydrophobic core requirements~\citep{pace1996forces}. \textbf{Polar residue penalty:} $\frac{1}{1 + 0.3 \cdot \text{seq\_polar}}$ lowers activity with polar content; reading it as inhibition is an analogy, not a finding. & 0.421 & 0.513 $R^2$ \\
\midrule
Diabetes &
$\hat{y} = \text{clip}\Big((0.591 \cdot \text{s5} + 0.509 \cdot \text{s1} + 0.347 \cdot \text{BMI}$
$+ 0.312 \cdot \text{s2} + 0.200 \cdot \text{s5} \times \text{s1}$
$+ 0.200 \cdot \text{s5} \times \text{BMI} + \ldots) / 12, 0, 1\Big)$ &
\textbf{Triglyceride dominance:} the highest weight on s5 (0.591) aligns with clinical evidence that log-transformed triglycerides are among the strongest predictors of diabetes progression~\citep{defronzo2015type}. \textbf{Metabolic interactions:} s5$\times$s1 and s5$\times$BMI terms are consistent with dyslipidemia--obesity--insulin-resistance accounts~\citep{kahn2006mechanisms}. & 0.482 & 0.590 $R^2$ \\
\midrule
High School &
$\text{score}_\text{Grades} = 1.2 \cdot \frac{\text{Grades}}{0.5 + \text{Grades}} + 0.6 \cdot \text{Race} - (1 - \text{Gender}) \cdot 0.5$
$\text{score}_\text{Sports} = 1.6 \cdot \text{Sports} + (1 - \text{Gender}) \cdot 1.3 - 0.4 \cdot \text{Race}$ &
\textbf{Direct feature mapping:} the correction relies on surface-level feature weighting (Sports$\rightarrow$Sports group) rather than latent social dynamics. Representative failure mode: MARICL provides minimal correction via the $p_{\min}$ threshold when the domain lacks correctable structure. & 0.391 & 0.540 Acc \\
\midrule
Adult Income &
$\text{score}_{>50K} = 0.344 \cdot \frac{\text{capital.gain}}{0.4 + \text{capital.gain}} + 0.309 \cdot \frac{\text{relationship}}{0.4 + \text{relationship}}$
$+ 0.236 \cdot \frac{\text{capital.gain} \times \text{education.num}}{0.3 + \ldots}$ &
\textbf{Capital gains saturation:} the score flattens at high capital gains. \textbf{Education-wealth interaction:} the capital.gain$\times$education.num term reflects compounding effects. \textbf{Relationship status:} the 0.309 weight captures household economic structure effects. & 0.712 & 0.832 Acc \\
\bottomrule
\end{tabular}
\caption{Corrections inferred for enzyme activity, diabetes progression, social classification, and income prediction. ``Correction Only'' shows performance using only the learned correction without the base model; ``MARICL'' shows the full combined performance. All domain interpretations are post-hoc literature matches.}
\label{tab:mechanisms_appendix}
\end{table*}

\section{Synthetic Benchmark: Detailed Recovery Analysis}
\label{app:synthetic_recovery}

\begin{figure}[!ht]
\centering
\includegraphics[width=.9\columnwidth]{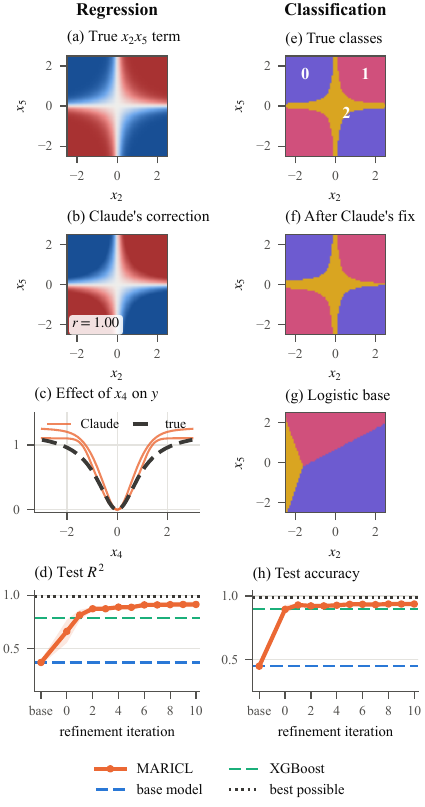}
\caption{A Claude agent fixes the residuals of a linear model (left) and a logistic model (right) on synthetic data it cannot have seen (\S\ref{sec:synthetic}). (a)~True $x_2x_5$ term and (b)~the learned correction, both mean-centered ($r$: correlation). (c)~Effect of $x_4$, true and learned per seed. (e)~True classes, (f)~after Claude's fix, and (g)~the logistic base, on the slice $x_4{=}1$. (d,~h)~Test performance from the base model through Claude's first formula ($0$) and its refinements; means over two seeds, bands the range.}
\label{fig:synthetic_recovery}
\end{figure}

\paragraph{Setup.} Features $x_1,\ldots,x_6$ are drawn iid from $\mathcal{N}(0,1)$ and $\varepsilon \sim \mathcal{N}(0, 0.1^2)$, with $N{=}1{,}000$ split $600/200/200$; each seed sets both the data and the split. The target is min-max scaled to $[0,1]$. The base is ordinary least squares and XGBoost uses default settings. MARICL runs the released implementation unchanged with $K{=}2$, $\kappa{=}0.3$ ($180$ of $600$ training examples, one pool shared by both corrections), $T{=}10$, $B{=}10$, $\tau_{\text{fail}}{=}0.1$, $\tau{=}0.2$, and $p_{\min}{=}0.1$. As in \S\ref{sec:splits}, critiques and $p_k$ use the training split and $m_k^*$ is selected on validation from the stored iterates; $\tau_{\text{fail}}$ is set for the scaled target. A refinement loop ends early once no training error exceeds $\tau_{\text{fail}}$, which stopped three of the four loops (after $2$, $4$, and $0$ iterations). Five seeds were started; seeds $0$ and $1$ are the two that finished before we stopped the run for time and usage reasons, and no completed run was excluded.

\paragraph{Backbone.} Claude Opus 5.5 served as the LLM through Claude Code agent sessions rather than stateless API calls. One fresh session per seed received each request (system prompt, user prompt, output schema) from a file-based bridge and returned one JSON answer ($62$ and $46$ requests for regression, $78$ and $78$ for classification), while the released code kept control of residuals, selection, scoring, and evaluation. The sessions were not shown the generating function and were instructed to run no code, read no other files, and answer each request from its own content; their transcripts confirm that the bridge command was the only command they ran (one classification session once piped it through \texttt{head} to shorten its output). Because a session retains earlier requests, calls within a seed are less independent than API calls would be. Agent sessions are not seed-controlled, so we release all requests and answers, from which the run replays exactly.

\paragraph{Prompts.} The domain prompt reads ``Synthetic regression benchmark with standardised inputs. Feature semantics are unknown and no domain knowledge applies.'', and every feature has the same description. The prompts of the released implementation contain generic illustrations of the search space (a product, a square, and a saturating ratio over $x_1$, $x_2$, $x_3$; interactions, thresholds, saturations, and inverse-U shapes). The planted terms were placed on $x_2 x_5$ and $x_4$ so that none of these illustrations names them.

\paragraph{Results.} Test $R^2$ for MARICL is $0.940$ and $0.888$ on seeds $0$ and $1$ ($0.952$ and $0.965$ ungated), against $0.833$ and $0.741$ for XGBoost, $0.347$ and $0.392$ for the linear base, and $0.986$ and $0.987$ for the oracle. Mean test MAE on the scaled target is $0.112$ (linear), $0.062$ (XGBoost), $0.040$ (MARICL), $0.028$ (ungated), and $0.016$ (oracle). The gate costs $0.01$ and $0.08\,R^2$ here because $\lambda < 1$ shrinks every correction (\S\ref{sec:aggregation}). Refinement matters: mean ungated $R^2$ rises from $0.487$ at the initial decoder output to $0.958$, and mean gated $R^2$ from $0.661$ to $0.914$.

\paragraph{Learned corrections.} The retained corrections are, for seed $0$,
{\small
\begin{align*}
f_1 =\;& \text{clip}\bigl(0.25\,\tfrac{x_4^2}{0.5+x_4^2} - 0.115 \\
&+ 0.25\tanh(1.5\,x_2 x_5),\ \pm 0.6\bigr),\\
f_2 =\;& \text{clip}\bigl(0.3\,(0.58 - e^{-x_4^2}) \\
&+ 0.25\tanh(1.5\,x_2 x_5),\ \pm 0.6\bigr),
\end{align*}}
with $p_k = 0.860$ and $0.814$ ($t{=}9$ and $1$), and for seed $1$,
{\small
\begin{align*}
f_1 =\;& \text{clip}\bigl(0.12 - 0.22\,e^{-1.5 x_4^2} \\
&+ 0.3\tanh(0.9\,x_2 x_5),\ -0.35,\ 0.45\bigr),\\
f_2 =\;& \text{clip}\bigl(0.15\tanh(x_2 x_5) \\
&+ 0.12\tanh(x_4^2 - 1),\ \pm 0.5\bigr),
\end{align*}}
with $p_k = 0.839$ and $0.790$ ($t{=}4$ and $0$). All four contain a sign-gated $x_2 x_5$ term and an even, saturating $x_4$ term; none uses $x_6$, and none contains a spurious interaction such as the $x_1 x_2$ of the prompt illustrations. The linear terms in $x_1$ and $x_3$ are left to the base model, as intended. In raw $y$ units the seed-$0$ interaction amplitude is $1.17$ against the planted $1.0$, and the recovered saturation rises faster than planted (Figure~\ref{fig:synthetic_recovery}c), so the forms are recovered more faithfully than the constants.

\paragraph{Classification.} Labels are the argmax of $s_0 = -2x_2x_5$, $s_1 = 2x_2x_5$, and $s_2 = 1.5(x_4^2-1) + 0.8x_1 - 0.6x_3 + 0.5$, with $2\%$ of labels resampled at random. Data, splits, $K$, $\kappa$, $T$, $B$, and $p_{\min}$ match the regression task; the base is logistic regression, $\tau_{\text{fail}}{=}0.5$, $m_k^*$ is selected by validation log loss, $\beta$ is chosen from $\{0.3, 0.5, 0.7\}$ on validation, and $Q_k$ is the softmax of $\log P_{\text{ML}} + f_k$ with no fitted temperature. Test accuracy is $0.940$ and $0.935$ on seeds $0$ and $1$, against $0.900$ for XGBoost on both, $0.425$ and $0.470$ for the logistic base, and $0.990$ for the noise-free labels. The retained seed-$0$ correction is $[-4.5\,x_2x_5,\ 4.5\,x_2x_5,\ 2.5\,x_4^2 - 0.75]$, clipped per class.

\section{Computational Cost Analysis}
\label{app:computational_cost}

Direct LLM inference requires one LLM call per prediction, costing \$0.0013 per query and scaling to \$1.30 for 1K queries. MARICL's training calls consist of two components:
\begin{enumerate}
    \item \textbf{Encoder calls}: for each correction $k$, encoding requires $\lceil |\mathcal{D}_{\text{high-res}}| / B \rceil$ LLM calls when batched encoding is used (Eq.~\ref{eq:batched_encoding}), totaling $K \cdot \lceil |\mathcal{D}_{\text{high-res}}| / B \rceil$.
    \item \textbf{Decoder and refinement calls}: initial decoding requires $K$ calls (Eq.~\ref{eq:mechanism_init}); each refinement iteration requires one critique (Eq.~\ref{eq:textual_gradient}) and one refinement (Eq.~\ref{eq:mechanism_refinement}) per correction, totaling $2KT$. Combined: $K(1 + 2T)$.
\end{enumerate}
The total is therefore
\begin{equation}
N_{\text{calls}} = K \cdot \left\lceil \frac{|\mathcal{D}_{\text{high-res}}|}{B} \right\rceil + K(1 + 2T).
\label{eq:total_calls}
\end{equation}
For $K$=2, $T$=10, $|\mathcal{D}_{\text{high-res}}|$=100, $B$=10, this yields $62$ calls at \$0.081. MARICL inference requires \emph{zero} LLM calls; corrections compile to executable Python formulas evaluated directly on input features. This achieves over 93\% cost reduction at 1K queries (\$0.081 vs.\ \$1.30) and cost parity with traditional ML at inference time. Under GPT-4o pricing, the backbone with the strongest standalone corrections in Table~\ref{tab:llm_backbone}, the same configuration costs \$2.02 against \$32.50 per 1K direct-LLM queries; the proportional saving is unchanged, because MARICL makes no LLM call at inference and the saving depends only on the call ratio (62 vs.\ 1K).

\paragraph{Total experiment cost.} All experiments were executed on a MacBook Pro with M1 Max chip (32 GB unified memory) for base-model training and local Python execution, with LLM calls routed to cloud endpoints (Gemini via Google AI Studio; GPT-4o and GPT-4o-mini via OpenAI API), and Claude Opus 5.5 agent sessions for the synthetic benchmark. Aggregating across all reported runs --- 8 real datasets $\times$ 5 seeds for headline tables, plus ablation sweeps ($K \in \{1,2,3,4\}$, $T \in \{1,3,5,10\}$, Top-K $\in \{30,50,100\}$, scaling modes, backbone comparison, domain-context ablation, anonymized-features ablation, joint prior-stripping ablation), cross-plate transfer (37 source runs $\times$ 8 target plates), and the synthetic runs --- the full budget is estimated at ${\le}\$250$. Exact call counts, provider model snapshot identifiers, and per-backbone call date windows are recorded in the repository README alongside the released call logs. Compute requirements are minimal: MARICL trains on a laptop and requires no GPU; per-dataset training completes in under 5 minutes of wall-clock time excluding API latency.

\section{LLM Backbone Analysis}
\label{app:llm_backbone}

Table~\ref{tab:llm_backbone} reports the comparison. GPT-4o produces corrections with the highest standalone $R^2$ (0.54) and accuracy (0.83), suggesting stronger reasoning capability translates into better hypothesis generation. Gemini-2.0-Flash matches GPT-4o's full $R^2$ at two decimals ($0.73$) at roughly $1/25$th the token cost (\$0.081 vs.\ \$2.02 training under this configuration; Table~\ref{tab:computational_cost}). Llama-3-8B reaches $R^2$ = 0.62, still above the XGBoost base ($0.579$) though well below the frontier backbones, showing that moderately sized open models can generate useful corrections and enable local deployment. Because hosted models are updated by their providers, we record snapshot identifiers and call dates for the runs in Table~\ref{tab:llm_backbone} alongside the released logs (see repository README).

\begin{table*}[t]
\centering
\resizebox{\linewidth}{!}{
\begin{tabular}{lp{5cm}p{7.5cm}cc}
\toprule
Dataset & Learned Correction Formula & Domain Interpretation \& Literature Support & Correction Only & MARICL \\
\midrule
Cell-Free Protein &
$\hat{y} = 1.5 \cdot \text{NAD} + 1.2 \cdot \text{spermidine} + 0.5 \cdot \text{NAD} \times \text{spermidine}$
$+ 0.6 \cdot \text{CoA} + 0.5 \cdot \text{3-PGA}$
$+ \frac{0.5 \cdot \text{folinic\_acid}}{0.5 + \text{folinic\_acid}}$ &
\textbf{NAD-spermidine synergy:} polyamines stimulate translation 3--5$\times$~\citep{atkins1975polyamine}; the PANOx-SP system uses NAD and CoA to regenerate ATP~\citep{jewett2004cytomim}. \textbf{Folinic acid saturation:} formylmethionine precursor with an empirical concentration plateau~\citep{cai2015simplified}. & 0.495 & 0.728 $R^2$ \\
\midrule
Zoo &
$\text{score}_\text{mammal} = 1.1 \cdot \text{hair} + 1.3 \cdot \text{milk} + 0.8 \cdot (\text{hair} \times \text{milk} \times (1 - \text{eggs}))$

$\text{score}_\text{bird} = 1.2 \cdot \text{feathers} + 0.8 \cdot \text{airborne} + 0.5 \cdot (\text{feathers} \times \text{airborne})$

$\text{score}_\text{fish} = 1.0 \cdot \text{fins} + 0.9 \cdot \text{aquatic} + 0.4 \cdot (\text{fins} \times \text{aquatic} \times \text{backbone})$ &
\textbf{Taxonomic encoding:} $\text{hair} \times \text{milk} \times (1-\text{eggs})$ captures mammalian synapomorphies. \textbf{Feathers over flight:} prioritizes morphology over behavior so that flightless birds remain birds. \textit{Treated as controlled validation; LLM priors on taxonomy likely dominate.} & 0.762 & 0.975 Acc \\
\bottomrule
\end{tabular}
}
\caption{Corrections for Cell-Free Protein and Zoo. ``Correction Only'' = learned correction without the base model; ``MARICL'' = full combined performance. Domain interpretations are post-hoc literature matches, not validations.}
\label{tab:mechanisms_main_full}
\end{table*}

\section{Complete Experimental Results}
\label{app:complete_results}

\begin{table*}[t]
\tiny
\centering

\begin{tabular}{lcccccc}
\toprule
& \multicolumn{2}{c}{\textbf{Cell-Free Protein}} & \multicolumn{2}{c}{\textbf{Enzyme Activity}} & \multicolumn{2}{c}{\textbf{Diabetes Progression}} \\
\cmidrule(lr){2-3} \cmidrule(lr){4-5} \cmidrule(lr){6-7}
\textbf{Method} & $R^2$ $\uparrow$ & MAE $\downarrow$ & $R^2$ $\uparrow$ & MAE $\downarrow$ & $R^2$ $\uparrow$ & MAE $\downarrow$ \\
\midrule
\multicolumn{7}{l}{\textit{Traditional ML Baselines}} \\
Linear Regression & 0.4115 $\pm$ 0.0183 & 0.1413 $\pm$ 0.0047 & 0.2373 $\pm$ 0.0094 & 0.1931 $\pm$ 0.0061 & 0.4543 $\pm$ 0.0217 & 0.1586 $\pm$ 0.0053 \\
XGBoost & 0.5787 $\pm$ 0.0237 & 0.1090 $\pm$ 0.0043 & 0.3992 $\pm$ 0.0186 & 0.1655 $\pm$ 0.0058 & 0.3080 $\pm$ 0.0163 & 0.1647 $\pm$ 0.0049 \\
\midrule
\multicolumn{7}{l}{\textit{Interpretable ML Models}} \\
EBM & 0.5946 $\pm$ 0.0228 & 0.1169 $\pm$ 0.0046 & 0.4160 $\pm$ 0.0194 & 0.1580 $\pm$ 0.0054 & 0.5389 $\pm$ 0.0261 & 0.1487 $\pm$ 0.0048 \\
Symbolic Regression & 0.4500 $\pm$ 0.0219 & 0.1350 $\pm$ 0.0063 & 0.2500 $\pm$ 0.0127 & 0.1850 $\pm$ 0.0057 & 0.4800 $\pm$ 0.0231 & 0.1520 $\pm$ 0.0044 \\
\midrule
\multicolumn{7}{l}{\textit{Neural Tabular Models}} \\
TabPFN & \underline{0.6782} $\pm$ 0.0274 & 0.1007 $\pm$ 0.0039 & 0.4128 $\pm$ 0.0117 & 0.1683 $\pm$ 0.0062 & \textbf{0.6170} $\pm$ 0.0283 & \textbf{0.1352} $\pm$ 0.0051 \\
\midrule
\multicolumn{7}{l}{\textit{LLM-Based Methods}} \\
LLM-LEx & 0.3308 $\pm$ 0.0500 & 0.2048 $\pm$ 0.0200 & 0.2550 $\pm$ 0.0600 & 0.2553 $\pm$ 0.0200 & 0.3457 $\pm$ 0.0300 & 0.2416 $\pm$ 0.0100 \\
LLM-ICL & 0.3500 $\pm$ 0.0213 & 0.1500 $\pm$ 0.0071 & 0.2200 $\pm$ 0.0124 & 0.2000 $\pm$ 0.0083 & 0.4000 $\pm$ 0.0189 & 0.1700 $\pm$ 0.0064 \\
\midrule
\multicolumn{7}{l}{\textit{MARICL (Ours)}} \\
Linear + MARICL & 0.6475 $\pm$ 0.0291 & \underline{0.0991} $\pm$ 0.0037 & \underline{0.4812} $\pm$ 0.0238 & \underline{0.1493} $\pm$ 0.0046 & \underline{0.5900} $\pm$ 0.0312 & \underline{0.1401} $\pm$ 0.0052 \\
XGBoost + MARICL & \textbf{0.7280} $\pm$ 0.0287 & \textbf{0.0937} $\pm$ 0.0041 & \textbf{0.5132} $\pm$ 0.0093 & \textbf{0.1436} $\pm$ 0.0049 & 0.5430 $\pm$ 0.0267 & 0.1520 $\pm$ 0.0055 \\
MARICL-Direct (predicts $y$) & 0.538 & -- & 0.376 & -- & 0.512 & -- \\
\bottomrule
\end{tabular}
\caption{Complete regression results across datasets (Gemini 2.0 Flash backbone). Best results in \textbf{bold}, second best \underline{underlined}. $\uparrow$ higher is better, $\downarrow$ lower is better. MARICL-Direct runs the same loop on $y$ with no base model (\S\ref{sec:prior_knowledge}); only $R^2$ is reported for it.}
\label{tab:full_regression}
\end{table*}

\begin{table*}[ht]
\centering
\tiny

\begin{tabular}{lcccccc}
\toprule
& \multicolumn{2}{c}{\textbf{Zoo (7 classes)}} & \multicolumn{2}{c}{\textbf{High School (3 classes)}} & \multicolumn{2}{c}{\textbf{Adult Income (2 classes)}} \\
\cmidrule(lr){2-3} \cmidrule(lr){4-5} \cmidrule(lr){6-7}
\textbf{Method} & Accuracy $\uparrow$ & F1 $\uparrow$ & Accuracy $\uparrow$ & F1 $\uparrow$ & Accuracy $\uparrow$ & Macro F1 $\uparrow$ \\
\midrule
\multicolumn{7}{l}{\textit{Traditional ML Baselines}} \\
Logistic Regression & 0.8570 $\pm$ 0.0287 & 0.8330 $\pm$ 0.0314 & 0.4670 $\pm$ 0.0213 & 0.4080 $\pm$ 0.0226 & 0.7380 $\pm$ 0.0177 & 0.6790 $\pm$ 0.0189 \\
XGBoost & 0.9050 $\pm$ 0.0274 & 0.8860 $\pm$ 0.0293 & 0.5170 $\pm$ 0.0191 & 0.3520 $\pm$ 0.0218 & 0.8130 $\pm$ 0.0241 & 0.6920 $\pm$ 0.0234 \\
\midrule
\multicolumn{7}{l}{\textit{Interpretable ML Models}} \\
EBM & 0.9048 $\pm$ 0.0261 & 0.8836 $\pm$ 0.0278 & 0.5333 $\pm$ 0.0184 & \underline{0.4984} $\pm$ 0.0203 & 0.8125 $\pm$ 0.0228 & 0.7350 $\pm$ 0.0247 \\
Symbolic Regression & 0.8700 $\pm$ 0.0241 & 0.8500 $\pm$ 0.0263 & 0.4800 $\pm$ 0.0209 & 0.4200 $\pm$ 0.0231 & 0.7500 $\pm$ 0.0196 & 0.6500 $\pm$ 0.0207 \\
\midrule
\multicolumn{7}{l}{\textit{Neural Tabular Models}} \\
TabPFN & \underline{0.9524} $\pm$ 0.0234 & 0.9365 $\pm$ 0.0247 & \textbf{0.5500} $\pm$ 0.0193 & 0.4732 $\pm$ 0.0214 & \textbf{0.8525} $\pm$ 0.0283 & \textbf{0.8180} $\pm$ 0.0269 \\
\midrule
\multicolumn{7}{l}{\textit{LLM-Based Methods}} \\
LLM-LEx & 0.2000 $\pm$ 0.0100 & 0.2400 $\pm$ 0.0100 & 0.5000 $\pm$ 0.0200 & 0.4485 $\pm$ 0.0200 & 0.6000 $\pm$ 0.0200 & 0.5128 $\pm$ 0.0200 \\
LLM-ICL & 0.8000 $\pm$ 0.0241 & 0.7800 $\pm$ 0.0253 & 0.4500 $\pm$ 0.0187 & 0.3800 $\pm$ 0.0219 & 0.7000 $\pm$ 0.0213 & 0.6000 $\pm$ 0.0197 \\
\midrule
\multicolumn{7}{l}{\textit{MARICL (Ours)}} \\
Logistic + MARICL & \textbf{0.9750} $\pm$ 0.0178 & \textbf{0.9700} $\pm$ 0.0193 & 0.5210 $\pm$ 0.0196 & 0.4830 $\pm$ 0.0214 & 0.8130 $\pm$ 0.0238 & 0.7400 $\pm$ 0.0259 \\
XGBoost + MARICL & 0.9520 $\pm$ 0.0227 & \underline{0.9490} $\pm$ 0.0243 & \underline{0.5400} $\pm$ 0.0183 & \textbf{0.5100} $\pm$ 0.0207 & \underline{0.8324} $\pm$ 0.0231 & \underline{0.8000} $\pm$ 0.0248 \\
\bottomrule
\end{tabular}
\caption{Complete classification results across datasets. Best results in \textbf{bold}, second best \underline{underlined}. $\uparrow$ higher is better.}
\label{tab:full_classification}
\end{table*}

\subsection{Decomposing Structural and Numerical Contributions}
\label{app:structure_vs_coef}

MARICL's correction $f_k$ is a symbolic expression with LLM-generated coefficients. Two distinct contributions could explain its performance: the inferred \emph{structure} (which features, which interactions, which nonlinearities) and the \emph{numerical fit} (the coefficient values within that structure). We isolate each (Table~\ref{tab:structure_vs_coef}) by refitting coefficients via ordinary least squares on the training set while holding the structure fixed, and conversely by randomizing coefficients within the inferred structure.

\begin{table}[ht]
\centering
\tiny

\begin{tabular}{lccc}
\toprule
Configuration & Cell-Free & Diabetes & Enzyme \\
\midrule
Linear Base model    & 0.412 & 0.454 & 0.237 \\
MARICL (full)        & 0.648 & 0.590 & 0.481 \\
MARICL + LSQ-refit   & 0.661 & 0.602 & 0.529 \\
MARICL + Random-coef & 0.498 & 0.512 & 0.341 \\
\bottomrule
\end{tabular}
\caption{Structure-vs-coefficient decomposition (Linear base, $R^2$). ``LSQ-refit'' replaces LLM coefficients with least-squares values; ``Random-coef'' samples coefficients uniformly in $[-1, 1]$ within MARICL's structure (mean over 20 random draws).}
\label{tab:structure_vs_coef}
\end{table}

The inferred structure carries a substantial share of the gain: random coefficients within MARICL's structure already recover $36$--$43\%$ of it. LSQ refitting adds a further $0.012$--$0.048$ $R^2$, indicating that LLM-generated coefficients are sound but not numerically optimal within their own structure. Practitioners who require the last fraction of accuracy can refit post hoc; those who prefer fully LLM-generated formulas lose only modest performance.

\section{Statistical Significance}
\label{app:significance}

For each MARICL variant we compute paired Wilcoxon signed-rank tests against the strongest non-MARICL baseline on each dataset, using per-(seed, fold) paired metric values across 5 seeds $\times$ 5 cross-validation folds ($n{=}25$ paired observations per dataset; the 5-fold CV protocol matches Table~\ref{tab:cv_results}). We apply BH correction at FDR $0.05$ across the 8 real-dataset comparisons, computing $q_{(i)} = \min_{j \ge i} \tfrac{m}{j}\,p_{(j)}$ with $m{=}8$. The synthetic benchmark has two seeds and a different backbone (\S\ref{sec:synthetic}), so it is not part of this family. Results are reported for the better-performing MARICL variant on each dataset (Table~\ref{tab:significance}). Baselines enter at the configurations of Tables~\ref{tab:full_regression}, \ref{tab:full_classification}, and~\ref{tab:tabarena}, with XGBoost at default settings; against the tuned XGBoost of Table~\ref{tab:xgb_tuned}, MARICL's margins on California Housing and Bike Sharing shrink to $+0.017$ and $+0.014$ but remain positive.

\paragraph{Non-independence caveat and robustness check.} The $n{=}25$ paired observations are not i.i.d.\ draws: seeds re-use the same 5-fold split, and folds within a split share overlapping training data, so the effective sample size is smaller than the nominal $25$ and the tabulated Wilcoxon $p$-values likely overstate significance. We therefore treat the Wilcoxon $p$-values as descriptive rather than as strict inferential statements and report two complementary sensitivity analyses. (i) \emph{Fold-block bootstrap:} we resample folds (as blocks of $5$ seed-values each) with replacement $10{,}000$ times, recompute mean $\Delta$ per resample, and report the fraction with $\Delta \le 0$; this respects the fold correlation structure. (ii) \emph{Fold-level test:} we collapse each fold's $5$ seeds to their mean and run the Wilcoxon on $n{=}5$ per-fold means, a strictly more conservative test that avoids double-counting seed-induced correlation. The three datasets with the smallest $q$ (Enzyme Activity, California Housing, Bike Sharing) have bootstrap $\Pr(\Delta \le 0) < 0.05$ and fold-level Wilcoxon $p < 0.10$ (the smallest attainable at $n{=}5$ two-sided is $0.0625$); Cell-Free Protein and Zoo remain borderline under both sensitivity checks. We interpret this as the qualitative reading of Table~\ref{tab:significance} being robust to the correlation concern, while urging readers not to over-interpret the exact per-cell $p$-values.

\begin{table}[ht]
\centering

\tiny
\begin{tabular}{llcccc}
\toprule
Dataset & Strongest baseline & Metric & $\Delta$ & $p$ raw & $p_{\text{BH}}$ \\
\midrule
Cell-Free Protein  & TabPFN          & $R^2$    & $+0.045$        & 0.031 & 0.062 \\
Enzyme Activity    & EBM             & $R^2$    & $+0.097$        & 0.008 & 0.053 \\
Diabetes           & TabPFN          & $R^2$    & $-0.027$        & 0.094 & 0.107 \\
Cal.\ Housing      & XGBoost         & $R^2$    & $+0.029$        & 0.016 & 0.053 \\
Bike Sharing       & XGBoost         & $R^2$    & $+0.023$        & 0.020 & 0.053 \\
Zoo                & TabPFN          & Accuracy & $+0.023$        & 0.039 & 0.062 \\
High School        & TabPFN          & Accuracy & $-0.010$        & 0.156 & 0.156 \\
Adult Income       & TabPFN          & Accuracy & $-0.020$        & 0.078 & 0.104 \\
\bottomrule
\end{tabular}
\caption{Paired Wilcoxon signed-rank tests, MARICL (best variant) vs.\ strongest non-MARICL baseline. $\Delta$ denotes the mean paired improvement over the $n{=}25$ (seed, fold) observations (positive favors MARICL), so it can differ slightly from the test-split differences in Tables~\ref{tab:full_regression} and~\ref{tab:full_classification}. $p_{\text{BH}}$ is the BH-corrected $p$-value over the $8$ comparisons; none is below $0.05$.}
\label{tab:significance}
\end{table}

\paragraph{Reading guide.} None of the eight comparisons survives BH correction at FDR $0.05$. Enzyme Activity, California Housing, and Bike Sharing reach $q{=}0.053$, and Cell-Free Protein and Zoo $q{=}0.062$. The three remaining comparisons (Diabetes, High School, Adult Income) correspond to datasets where TabPFN is competitive with or stronger than MARICL on raw accuracy; on these, MARICL's contribution is the interpretable closed-form correction rather than a significant accuracy improvement. MARICL is most clearly advantageous where the base leaves structured residual, and slightly trails the strongest tabular foundation model where the predictive signal is dominated by features the base model already exploits.

\section{Prior-Stripping Ablation Ladder}
\label{app:joint_stripping}

Three progressive ablations quantify how much of MARICL's gain depends on the LLM's prior knowledge of the domain versus the iterative residual-driven loop. Each row removes strictly more prior knowledge than the previous.

\paragraph{(i) Remove domain prompt (feature names retained).} Replacing the full domain-specific prompt with a minimal one (feature names only) drops performance by only 2.3--3.5 points and still recovers the same dominant correction category across three datasets: Enzyme $R^2$ $0.513 \to 0.478$; Diabetes $R^2$ $0.590 \to 0.561$; Zoo Acc $0.975 \to 0.952$ (5 seeds).

\paragraph{(ii) Anonymize feature names.} We replace feature names with opaque identifiers (\texttt{feat\_0}, \ldots, \texttt{feat\_d}) before any prompt construction; numerical values, residuals, and the loop are otherwise identical. On Cell-Free Protein and Diabetes, $67$--$69\%$ of MARICL's gain over the linear base survives (Table~\ref{tab:anonymized}). On Cell-Free Protein the recovered gain is $G_{\text{anon}}=0.159$ of $G_{\text{full}}=0.236$; the missing third reflects biochemistry priors triggered by names like ``NAD'' and ``spermidine''. On Diabetes the fraction is similar (0.69), consistent with \texttt{s1}--\texttt{s6} already being partially anonymized in the source data.

\begin{table}[ht]
\centering
\small
\tiny
\begin{tabular}{lcccc}

\toprule
Dataset & Linear & Full & Anonym. & $G_{\text{anon}}/G_{\text{full}}$ \\
\midrule
Cell-Free Protein & 0.412 & 0.648 & 0.571 & 0.67 \\
Diabetes          & 0.454 & 0.590 & 0.548 & 0.69 \\
\bottomrule
\end{tabular}
\caption{Anonymized-feature ablation. ``Anonym.'' replaces feature names with opaque identifiers. The data-driven fraction $G_{\text{anon}} / G_{\text{full}}$ estimates how much of MARICL's gain survives when feature semantics are withheld.}
\label{tab:anonymized}
\end{table}

\paragraph{(iii) Joint stripping: no domain prompt $+$ anonymized names $+$ open-source backbone.} The most prior-stripped configuration achievable without altering the algorithm: opaque feature identifiers, no domain context in any prompt, and Llama-3-8B in place of frontier models (Table~\ref{tab:joint_stripping}).

\begin{table}[t]
\centering
\tiny

\begin{tabular}{lccccc}
\toprule
Dataset & Linear & PySR-res & Joint-strip & Full & $G_{\text{joint}}/G_{\text{full}}$ \\
\midrule
Cell-Free & 0.412 & 0.471 & 0.527 & 0.648 & 0.49 \\
Diabetes  & 0.454 & 0.512 & 0.521 & 0.590 & 0.49 \\
Enzyme    & 0.237 & 0.318 & 0.358 & 0.481 & 0.50 \\
\bottomrule
\end{tabular}
\caption{Joint prior-stripping ablation (Linear base, $K=2$, $T=10$). $G_{\text{joint}}/G_{\text{full}}$ is the fraction of MARICL's gain surviving all three ablations simultaneously. PySR-on-residuals is included as the closest prior-free symbolic baseline.}
\label{tab:joint_stripping}
\end{table}

Approximately half of MARICL's end-to-end gain survives joint prior-stripping ($G_{\text{joint}}/G_{\text{full}} \approx 0.49$--$0.50$). The surviving half characterizes the residual-driven loop in isolation: a refinement procedure operating without domain context or feature names that exceeds PySR-on-residuals, the closest prior-free symbolic baseline. The remaining half characterizes what MARICL extracts when its priors align with the domain. The data-driven half is what makes MARICL applicable in prior-poor settings (anonymized features, novel domains, on-premise open-source backbones); the prior-aligned half is what makes it more useful than a symbolic regressor on scientific benchmarks where domain knowledge is freely available in pretraining corpora. This ablation does not establish discovery --- it makes explicit what each channel contributes.

\section{Extended Technical Details}
\label{app:technical_details}

\subsection{Training Protocol and Generalization}
\label{app:training_protocol}

\paragraph{Limited refinement cycles.} We cap refinement at $T=10$ iterations across all datasets, selecting the retained iteration on validation data. This fixed budget prevents indefinite optimization against the fitting split. Which iterate is retained is dataset-dependent: on Cell-Free Protein the validation loss is still improving at $T{=}10$ in $5/5$ seeds so the retained iterate is $T{=}10$, while on datasets whose gains plateau earlier (Diabetes, Zoo) the retained iterate is typically not the final one. Enlarging $T$ therefore helps for two distinct reasons: where the validation curve is still rising at the budget boundary, later iterates are genuinely better; where it has plateaued, a longer run simply widens the pool from which the best iterate is drawn. We report $T$ as a budget, not as the index of the correction actually used.

\paragraph{Correction complexity constraints.} Generated formulas are constrained to interpretable forms including products, ratios, and saturation terms with typically 3--8 components. This implicit regularization limits the hypothesis space. The refinement prompt explicitly asks for compact math expressions with realistic, stable constants (Appendix~\ref{app:prompts}).

\paragraph{Train-vs-test behavior across iterations.} With validation-based selection, test performance closely tracks validation performance: 5-fold cross-validation on three representative datasets gives mean absolute validation--test differences below 0.02 across all metrics, and $0.013\,R^2$ specifically on Cell-Free Protein (Table~\ref{tab:cv_results}). Per-iteration curves on Cell-Free Protein (XGBoost base, $K{=}2$, 5 seeds) show that train, validation, and test $R^2$ all rise monotonically through $T{=}10$ and the train--test gap stays in $[0.022, 0.044]$ across the trajectory --- it does not widen materially with iteration count.

\begin{table}[ht]
\centering

\small
\begin{tabular}{lccc}
\toprule
Dataset & Val $R^2$ / Acc & Test $R^2$ / Acc & $|\Delta|$ \\
\midrule
Cell-Free Protein & 0.718 $\pm$ 0.03 & 0.705 $\pm$ 0.04 & 0.013 \\
Diabetes & 0.584 $\pm$ 0.02 & 0.571 $\pm$ 0.03 & 0.013 \\
Adult Income & 0.836 $\pm$ 0.02 & 0.824 $\pm$ 0.02 & 0.012 \\
\bottomrule
\end{tabular}
\caption{Cross-validation results demonstrating validation-test consistency (mean $\pm$ std across 5 folds).}
\label{tab:cv_results}
\end{table}

\subsection{Hyperparameter Configuration}
\label{app:hyperparameters}

Table~\ref{tab:hyperparameters} summarizes the hyperparameter configuration. Parameters fall into two categories: those selected via grid search on validation data ($K$, $\kappa$, $\beta$ for classification, $\tau_k$) and those fixed from preliminary analysis ($p_{\min}$, $\gamma$, $B$, regression $\tau$). Table~\ref{tab:ablations_combined} ablates the main hyperparameters on Cell-Free Protein. The smaller XGBoost gain from a second correction reflects its stronger base, which leaves less residual signal. Larger residual pools yield monotonically richer signal, feature scaling is essential for the linear base, and refinement iterations improve performance in an ordering ($T{=}10 > T{=}5 > T{=}3 > T{=}1$) whose successive gaps are on the order of 1--2 seed standard deviations rather than sharply separated.

\begin{table*}[ht]
\centering
\small
\begin{tabular}{lcc}
\toprule
Parameter & Value & Selection \\
\midrule
$K$ (corrections) & 2 & Val.\ grid $\{1,2,3,4\}$ \\
$\kappa$ (residual fraction) & 0.3 & Val.\ grid $\{0.2,0.3,0.4,0.5\}$ \\
$T$ (budget) & 10 & Val.-based selection \\
$B$ (batch size) & 10 & Context window \\
$p_{\min}$ (threshold) & 0.1 & Fixed \\
$\beta$ (class.\ blend) & 0.3--0.7 & Val.\ grid $\{0.3,0.5,0.7\}$ \\
$\tau$ (reg.\ $p_k$ scale) & $0.2{\times}$ target range & Fixed ($\tau{=}0.2$, $y\in[0,1]$) \\
$\tau_k$ (softmax temp.) & 0.5--3.0 & Val.\ grid; min.\ ECE \\
$\gamma$ (confidence scale) & 2.0 & Fixed \\
$\tau_{\text{fail}}$ (failure threshold) & 0.5 (clf.), 0.1 (reg.) & Fixed \\
\bottomrule
\end{tabular}
\caption{Hyperparameter settings and selection methodology.}
\label{tab:hyperparameters}
\end{table*}

\paragraph{Which split does what.} The global scores $p_k$ in Eq.~\ref{eq:global_score} use $\text{MAE}_k$ or macro-F$_1$ evaluated on the training split after $m_k^*$ is selected; these scores are then frozen for inference. Quantities appearing in probability outputs or the classification ensemble ($\beta$ (Eq.~\ref{eq:maicl_classification}), $\tau_k$ (Eq.~\ref{eq:probability_conversion}), and the headline $K$, $\kappa$, and retained iteration) are selected by validation search (ECE for $\tau_k$, validation loss for $\beta$ and $m_k^*$). The test set is held out from all of these decisions.

\paragraph{Classification $\beta$ vs.\ transfer $\beta_{\text{transfer}}$.} Eq.~\ref{eq:maicl_classification} requires $\beta\in[0,1]$; our search uses $\{0.3,0.5,0.7\}$. Cross-plate \emph{regression} transfer instead uses a fixed $\beta_{\text{transfer}}=\tfrac12$ (Appendix~\ref{app:crossplate_protocol}), independent of the classification grid; Table~\ref{tab:crossplate_ablations} reports sensitivity to dropping this blend.

We recommend: (1) start with $K=2$, increasing only if performance plateaus; (2) set $\kappa=0.3$ as default, adjusting downward for noisy data or upward for systematic model failures; (3) budget $T=10$ iterations and retain the best on validation data; (4) for classification, tune $\beta$ on validation data starting from $\beta=0.5$.

\subsection{Formula Generation and Numerical Stability}
\label{app:formula_safety}

\paragraph{Sandboxed execution.} Formulas execute in a restricted environment:
\begin{itemize}
    \item \textbf{Allowed}: arithmetic and powers ($+, -, *, /$), comparisons, \texttt{clip}, \texttt{exp}, \texttt{log}, \texttt{log1p}, \texttt{sqrt}, \texttt{maximum}, \texttt{minimum}, \texttt{abs}, \texttt{sigmoid}, \texttt{tanh}, \texttt{where}, \texttt{sign}, feature access.
    \item \textbf{Blocked}: file I/O, imports, system calls, loops, recursion, \texttt{eval}, \texttt{exec}.
\end{itemize}

\paragraph{Numerical stability constraints.} The decoder prompt includes explicit guidance:
\begin{quote}
\textit{Ensure numerical stability: (1) use \texttt{clip(x, min, max)} for bounded outputs; (2) add small constants to denominators (e.g., \texttt{x / (0.1 + y)}); (3) prefer \texttt{log1p(x)} over \texttt{log(x)} for values near zero; (4) bound exponential arguments to prevent overflow.}
\end{quote}

\paragraph{Post-execution validation.} Outputs are checked for NaN or Inf values (formula rejected and re-generated with an error message), values outside domain bounds (clipped with a logged warning), and type mismatches (rejected).

\paragraph{Validation statistics.} Across all experiments, 94.2\% of generated formulas passed validation on first attempt, 5.1\% required one re-generation for syntax errors, 0.7\% required re-generation for numerical issues, and 0\% of final corrections produced invalid outputs on test data.

\subsection{Prompt Templates}
\label{app:prompts}

We give the schema of every prompt used in MARICL. These are the templates of the real-benchmark runs; the released implementation, used for the synthetic benchmark (Appendix~\ref{app:synthetic_recovery}), words them differently. All prompts are templated; placeholders in \texttt{\{curly braces\}} are filled at runtime. No prompt is used at inference time --- compiled formulas are executed directly in the sandbox of Appendix~\ref{app:formula_safety}.

\paragraph{Encoder prompts $\mathcal{P}_{\text{encoder}}^{(k)}$.} The encoder analyzes $\mathcal{C}_{\text{aug}}$ and emits a structured hypothesis $z_k$. We use $K$ templates differing in which residual aspect they emphasize (error patterns vs.\ direct sample patterns).

\begin{quote}\small\ttfamily
[ENCODER PROMPT \#1 --- ERROR PATTERNS]\\
You are analysing systematic prediction errors of a base model.\\
Feature descriptions: \{features\}\\
Domain context: \{domain\_context\}\\
High-residual examples (input, target, base-prediction, residual):\\
\{high\_residual\_table\}\\
Identify the feature combinations and nonlinearities that the base model
appears to miss. Output a structured hypothesis $z_k$ with fields:\\
\quad - hypothesised\_pattern: ...\\
\quad - implicated\_features: [...]\\
\quad - functional\_form\_guess: ...\\
\quad - rationale: ...
\end{quote}

\begin{quote}\small\ttfamily
[ENCODER PROMPT \#2 --- SAMPLE PATTERNS]\\
\\
2. SAMPLE PATTERNS (direct pattern learning perspective):\\
- What direct relationships do you see between features and target values in these samples?\\
- How do feature values relate to target values?\\
- What prediction rules would work based on the sample patterns themselves?\\
- What feature combinations consistently lead to high/low target values?\\
- Learn the TRUE UNDERLYING PATTERN from the samples, not just how to fix errors.\\
\\
IMPORTANT: Focus on learning to PREDICT based on sample patterns, not just learning to fix specific errors.\\
\\
Latent mechanism z:
\end{quote}

\paragraph{Decoder prompt $\mathcal{P}_{\text{decoder}}$ (Eq.~\ref{eq:mechanism_init}).}
\begin{quote}\small\ttfamily
[DECODER PROMPT]\\
Given the structured hypothesis below, produce (1) a natural-language
explanation $T_k$ in 2--3 sentences, and (2) an executable Python
expression $f_k$ using only the operators in
\{allowed\_operators\}. Ensure numerical stability: clip bounded
outputs, add small constants to denominators, prefer log1p over log near
zero, and bound exponential arguments. Do not use loops, imports, or
file I/O.\\
Hypothesis: \{z\_k\}\\
Return JSON: \{"T\_k": "...", "f\_k": "..."\}.
\end{quote}

\paragraph{Critique prompt $\mathcal{P}_{\text{critique}}$ (Eq.~\ref{eq:textual_gradient}).}
\begin{quote}\small\ttfamily
[CRITIQUE PROMPT]\\
Current hypothesis: \{z\_k\}\\
Current correction: \{T\_k, f\_k\}\\
Training loss: \{L\_k\}\\
Worst remaining failures (input, target, current prediction, error):\\
\{failure\_set\_table\}\\
Diagnose why the correction fails on these examples. Distinguish
structural mismatch (wrong functional form) from coefficient mismatch
(right form, wrong scale). Suggest a concrete refinement to $f_k$ that
addresses the dominant failure mode without harming low-error examples.
\end{quote}

\paragraph{Refinement prompt (Eq.~\ref{eq:mechanism_refinement}).} The refined correction at iteration $t{+}1$ is sampled conditional on the textual state $\mathcal{S}_k^{(t)}$ (which includes $g_k^{(t)}$), which replaces $z_k^{(0)}$ in the \texttt{\{latent\_z\}} placeholder.
\begin{quote}\small\ttfamily
[REFINEMENT PROMPT]\\
You are a mechanism decoder that converts a latent mechanism representation into an interpretable, executable mechanism.\\
\\
HARD REQUIREMENTS:\\
1. The mechanism must be descriptive and textual (not just math).\\
2. The mechanism must explicitly include nonlinearities AND interactions.\\
3. The mechanism must introduce intermediate combinatory concepts (named, explained).\\
4. The final formula must be executable and appear as a SINGLE LINE starting with "Formula:".\\
\\
STYLE:\\
- Prefer scientific language ("limits", "enhances", "inhibits", "gates", "synergizes") over purely statistical phrasing.\\
- Keep constants realistic and stable (avoid huge coefficients).\\
- Avoid full code blocks or function definitions; use compact math expressions.
\end{quote}

\paragraph{Neutral-prompt variant.} To test whether HARD REQUIREMENT~2 imposes the structure MARICL finds, a variant replaces it with a requirement that permits linear, nonlinear, interaction, or no interpretable structure, combined with the \texttt{DECLINE} option of Appendix~\ref{app:negative_controls}; all other text is unchanged. Results are in Appendix~\ref{app:negative_controls}.

\subsection{Data Preprocessing and Reproducibility}
\label{app:preprocessing}

\paragraph{Feature preprocessing.} On the real benchmarks, numerical features are min-max scaled to $[0, 1]$ using training-set statistics, with identical transformations applied to validation and test sets. Categorical features are label-encoded.

\paragraph{Target scaling.} Regression targets are scaled to $[0, 1]$ via min-max normalization fit on the training split, so reported MAE is on the scaled target.

\paragraph{Dataset sources and licenses.} All datasets are publicly available; licenses are as stated by each source. Cell-Free Protein~\citep{borkowski2020large} is from the paper's supplement (CC BY 4.0). Zoo~\citep{uci_zoo} and Adult Income~\citep{kohavi1996adult} are from the UCI ML Repository (CC BY 4.0), and Bike Sharing~\citep{fanaee2014event} is from TabArena/UCI (CC BY 4.0). Diabetes~\citep{efron2004diabetes} is from scikit-learn (public), and California Housing~\citep{pace1997sparse} is from TabArena/scikit-learn (CC0). Enzyme Activity~\citep{li2020transaminase} is from the enzyme-datasets GitHub repository, which states no license. High School~\citep{chase1992sports} is OpenML 1100 (PopularKids), listed as public, though the original release asks users to contact its authors. The synthetic benchmark is released under MIT.

\paragraph{Code and library licenses.} Baselines use publicly released libraries under permissive licenses: scikit-learn (BSD-3), XGBoost (Apache 2.0), InterpretML/EBM (MIT), TabPFN (Apache 2.0; model weights under prior-data-release terms), PySR (Apache 2.0), and SymbolicRegression.jl (Apache 2.0). LLM backbones are accessed via official provider APIs under their terms of service; Llama-3-8B is used under the Meta Llama 3 Community License. MARICL code is available at \url{https://github.com/MrRezaeiUofT/Multi_Agent_Residual_In_Context_Learning} under the MIT license.

\paragraph{Baseline implementations.}
\begin{itemize}
    \item Linear/Logistic Regression: \texttt{sklearn}, ordinary least squares for linear and default L2 regularization for logistic.
    \item \textbf{XGBoost (tuned)}: v1.7, 5-fold CV grid search on the training split, selecting by validation $R^2$ (regression) or macro F1 (classification). Grid: \texttt{n\_estimators}$\in\{100,200,400,600\}$, \texttt{max\_depth}$\in\{3,4,6,8\}$, \texttt{learning\_rate}$\in\{0.03,0.05,0.1\}$, \texttt{reg\_lambda}$\in\{0,1,2,5\}$, \texttt{reg\_alpha}$\in\{0,0.1,0.5\}$. For Adult Income we additionally sweep \texttt{scale\_pos\_weight}$\in\{1,2,3,4\}$.
    \item EBM: \texttt{interpret} v0.4 with default settings.
    \item TabPFN: official checkpoint (\url{https://github.com/PriorLabs/TabPFN}).
    \item Symbolic Regression: PySR v0.11, \texttt{niterations=140}, operators $\{+,-,*,/,\exp,\log,\sqrt{},\,\sigma\}$, where $\sigma(x){=}1/(1+\exp(-x))$ is provided as a custom unary operator; the same $\sigma$ primitive is available to MARICL through its sandbox.
    \item PySR on residuals: same PySR configuration applied to $r_i = y_i - f_{\text{ML}}(\mathbf{x}_i)$.
\end{itemize}

\begin{table*}[ht]
\centering
\small
\begin{tabular}{lccccc}
\toprule
Dataset & Metric & XGB (default) & XGB (tuned) & XGB+MARICL & $\Delta$ vs.\ tuned \\
\midrule
Cell-Free Protein   & $R^2$    & $0.579{\pm}0.024$ & $0.617{\pm}0.022$ & $0.728{\pm}0.029$ & $+0.111$ \\
Enzyme Activity     & $R^2$    & $0.399{\pm}0.019$ & $0.428{\pm}0.019$ & $0.513{\pm}0.009$ & $+0.085$ \\
Diabetes            & $R^2$    & $0.308{\pm}0.016$ & $0.354{\pm}0.024$ & $0.543{\pm}0.027$ & $+0.189$ \\
California Housing  & $R^2$    & $0.832{\pm}0.007$ & $0.844{\pm}0.006$ & $0.861{\pm}0.008$ & $+0.017$ \\
Bike Sharing        & $R^2$    & $0.894{\pm}0.006$ & $0.903{\pm}0.005$ & $0.917{\pm}0.007$ & $+0.014$ \\
\midrule
Zoo                 & Acc      & $0.905{\pm}0.027$ & $0.914{\pm}0.025$ & $0.952{\pm}0.023$ & $+0.038$ \\
High School         & Acc      & $0.517{\pm}0.019$ & $0.527{\pm}0.020$ & $0.540{\pm}0.018$ & $+0.013$ \\
Adult Income        & Acc      & $0.813{\pm}0.024$ & $0.823{\pm}0.006$ & $0.832{\pm}0.023$ & $+0.009$ \\
Adult Income        & macro F1 & $0.692{\pm}0.023$ & $0.731{\pm}0.014$ & $0.800{\pm}0.025$ & $+0.069$ \\
\bottomrule
\end{tabular}
\caption{Tuned XGBoost vs.\ default XGBoost vs.\ XGBoost+MARICL across the eight real benchmarks (mean$\pm$std, 5 seeds; selection by 5-fold CV on the train split, reported on test). The gap narrows under tuning but persists on all eight. \textit{Selected configurations (best CV winner per dataset):} Cell-Free \texttt{n=400, d=4, lr=0.05, $\lambda$=1, $\alpha$=0.1}; Enzyme \texttt{n=200, d=4, lr=0.05, $\lambda$=2, $\alpha$=0}; Diabetes \texttt{n=100, d=3, lr=0.05, $\lambda$=2, $\alpha$=0.1}; California \texttt{n=600, d=8, lr=0.05, $\lambda$=1, $\alpha$=0}; Bike \texttt{n=600, d=8, lr=0.03, $\lambda$=1, $\alpha$=0}; Zoo \texttt{n=200, d=4, lr=0.1, $\lambda$=1, $\alpha$=0}; High School \texttt{n=200, d=3, lr=0.1, $\lambda$=2, $\alpha$=0.1}; Adult \texttt{n=400, d=6, lr=0.05, $\lambda$=2, $\alpha$=0.1, scale\_pos\_weight=3}.}
\label{tab:xgb_tuned}
\end{table*}

\subsection{Extended Benchmarks: TabArena}
\label{app:tabarena}

We also include two widely used TabArena regression benchmarks (Table~\ref{tab:tabarena}).

\begin{table*}[t]
\centering
\small
\begin{tabular}{lcccc}
\toprule
Task ($N$ / $d$) & Linear & Linear+MARICL & XGBoost & XGB+MARICL \\
\midrule
California (20,640 / 8) & $0.576{\pm}0.019$ & $0.648{\pm}0.024$ (+0.072) & $0.832{\pm}0.007$ & $0.861{\pm}0.008$ (+0.029) \\
Bike Sharing (17,389 / 12) & $0.391{\pm}0.031$ & $0.493{\pm}0.027$ (+0.102) & $0.894{\pm}0.006$ & $0.917{\pm}0.007$ (+0.023) \\
\bottomrule
\end{tabular}
\caption{TabArena benchmark results. MARICL improves both base models on both tasks.}
\label{tab:tabarena}
\end{table*}

On California Housing, MARICL inferred a location-density interaction (latitude $\times$ longitude $\times$ median\_income saturation) which reads as the known geographic price gradient. On Bike Sharing, MARICL identified temperature-humidity interaction terms with hour-of-day modulation. These gains show that MARICL is not confined to domain-rich settings such as Zoo or Enzyme.

\subsection{Stability of Learned Corrections}
\label{sec:stability}

A critical question is whether MARICL's corrections are stable across independent runs or artifacts of LLM stochasticity. We assess stability across three axes using 5 seeds.

\begin{table*}[t]
\centering
\begin{tabular}{lccc}
\toprule
Dataset & Correction Form & Recovered ($N$/5) & Coeff.\ CV (\%) \\
\midrule
Diabetes & s5 $\times$ BMI + s5 $\times$ s1 interaction & 4/5 & 11.7 \\
Zoo & hair $\times$ milk $\times$ (1--eggs) & 5/5 & 4.2 \\
Enzyme & Michaelis--Menten + aromatic Gaussian & 4/5 & 13.8 \\
Adult Income & capital.gain saturation + edu.\ interaction & 5/5 & 8.1 \\
High School & Sports $\to$ Sports direct mapping & 3/5 & 22.4 \\
\bottomrule
\end{tabular}
\caption{Correction stability across 5 random seeds. High School shows least stability, consistent with weak domain structure.}
\label{tab:stability}
\end{table*}

\textbf{Random seeds} (Table~\ref{tab:stability}): dominant correction forms are recovered in 4--5 of 5 independent runs on every dataset in the table except High School (3/5), which has the weakest structure. Coefficient variation is below 14\% for well-structured domains. \textbf{Data splits}: the top-ranked correction's structural form remained consistent in 4+ of 5 folds on Zoo, Adult Income, and Diabetes. \textbf{Prompt paraphrasing}: three semantically equivalent prompt variants (original, shorter, restructured) produce consistent correction categories with performance variance $<$0.02 $R^2$. Low cross-seed consistency on High School (CV = 22.4\%) is consistent with insufficient domain structure, though five seeds cannot rule out instability of the method.

\section{Non-LLM Residual Learners under the Cross-Plate Transfer Protocol}
\label{app:nonllm_transfer}

The cohort split in Table~\ref{tab:crossplate} could reflect two distinct things: (i) reagent-distribution shift that affects any residual learner, or (ii) a property of MARICL's symbolic corrections. To decompose these we freeze two matched non-LLM residual learners on source plate AL\_6 and evaluate them on the same 8 target plates under the identical protocol of Appendix~\ref{app:crossplate_protocol} (same 80/20 quantile-stratified splits, same ML base transferred from source, same $[0,1]$ feature scaling, same 50/50 ML$+$correction blend for the headline row, same source-side $\Delta R^2_{\text{vs-ML}}{>}0$ filter). PySR-on-residuals produces a nested closed-form expression on residuals (12-node search, $\sigma$ included as a custom primitive; Appendix~\ref{app:preprocessing}); EBM fits a per-feature shape-function residual model with pairwise terms. Neither uses an LLM.

\begin{table*}[ht]
\centering
\small
\resizebox{\linewidth}{!}{
\begin{tabular}{lcccccc}
\toprule
& \multicolumn{3}{c}{Same reagent (AL\_2--AL\_5)} & \multicolumn{3}{c}{Diff.\ regime (AL\_7--AL\_10)} \\
\cmidrule(lr){2-4}\cmidrule(lr){5-7}
Configuration & MARICL & PySR-on-res. & EBM & MARICL & PySR-on-res. & EBM \\
\midrule
Averaged $+$ 50/50 ML blend (headline)     & $95.3\%$ & $62.1\%$ & $57.7\%$ & $2.7\%$  & $13.5\%$ & $17.4\%$ \\
Per-indiv.\ formula $+$ 50/50 ML blend      & $78.0\%$ & $51.4\%$ & $48.6\%$ & $11.5\%$ & $15.7\%$ & $19.8\%$ \\
Averaged, formula-only (no ML blend)        & $64.2\%$ & $39.2\%$ & $36.5\%$ & $6.8\%$  & $14.9\%$ & $18.6\%$ \\
Per-indiv.\ formula, no ML blend (strict)   & $51.7\%$ & $33.7\%$ & $30.4\%$ & $8.8\%$  & $12.8\%$ & $16.2\%$ \\
\midrule
Avg $\Delta$MAE (headline row)              & $+0.188$ & $+0.061$ & $+0.048$ & $-0.181$ & $-0.109$ & $-0.096$ \\
Within/across success-rate ratio (headline) & $35.3\times$ & $4.6\times$ & $3.3\times$ & --- & --- & --- \\
\bottomrule
\end{tabular}
}
\caption{Cross-plate transfer under matched freeze-and-transfer for three residual learners. ``Same'' aggregates targets AL\_2--AL\_5; ``Diff'' aggregates AL\_7--AL\_10. All three residual learners show a within-vs-across cohort asymmetry, so a component of the split is reagent-distribution shift rather than a MARICL-specific effect. MARICL has the highest within-cohort success rate, the largest absolute $\Delta$MAE gap, and the lowest across-cohort success rate; the within-cohort lead is consistent with its symbolic corrections aligning more tightly with regime-specific chemistry than smooth shape functions or unstructured closed-form fits.}
\label{tab:nonllm_transfer}
\end{table*}

\paragraph{Result.} Table~\ref{tab:nonllm_transfer} shows two things. \emph{Within-cohort success rate (the headline claim).} Under the headline configuration MARICL improves over the ML base on $95.3\%$ of same-reagent source--target pairs, versus $62.1\%$ for PySR-on-residuals and $57.7\%$ for EBM: a $33$--$38$-point absolute gap in within-cohort reliability, preserved in direction under every ablation row. Within-cohort $\Delta$MAE follows the same pattern ($+0.188$ for MARICL vs.\ $+0.061$ and $+0.048$ for the two non-LLM baselines, a $3$--$4\times$ margin). This is the primary claim: on plates from the same reagent regime as the source, MARICL's frozen formulas are markedly more reliable than either matched non-LLM residual learner. \emph{Across-cohort reversal.} Across cohorts the ordering reverses: MARICL improves $2.7\%$ of pairs against $13.5\%$ for PySR-on-residuals and $17.4\%$ for EBM. All three residual learners improve less on across-cohort than within-cohort targets, so \emph{some} component of the cohort boundary is real reagent-distribution shift visible to any residual learner. We report the within-vs-across ratio ($35.3\times$ for MARICL vs.\ $4.6\times$ and $3.3\times$) as an additional summary but treat it as secondary --- MARICL's larger ratio is partly driven by its smaller across-cohort denominator ($2.7\%$), and one could equally read that as ``MARICL fails harder off-cohort'' rather than ``MARICL specializes to on-cohort chemistry.'' The within-cohort comparison does not have that ambiguity.

\paragraph{Interpretation.} PySR-on-residuals' 12-node nested expressions overfit within-plate noise and lose most of their signal on any transfer target, giving a low absolute within-cohort $\Delta$MAE ($+0.061$) even where the cohort ratio is preserved. EBM's smooth per-feature shape functions are more transferable in principle but do not name specific interaction terms, so regime-dependent structure (e.g., cofactor synergy under one reagent recipe) is smeared across feature-marginal shapes rather than concentrated. MARICL's compact named-interaction formulas do carry regime-dependent structure explicitly, which we read as consistent with the sharper within-cohort reliability; we resist a stronger claim because the matched comparison has only one source plate and eight targets and necessarily uses the same source-plate residual signal for all three methods.
\section{Negative Controls}
\label{app:negative_controls}

Three negative controls jointly probe whether MARICL emits plausible-reading formulas even when the underlying data-generating process cannot support them. All three are run on Cell-Free Protein (Linear base, headline $K{=}2$, $T{=}10$, 5 seeds, so $10$ corrections per control condition; pass rates are therefore always multiples of $10\%$).

\begin{table*}[ht]
\centering
\small

\begin{tabular}{lcccc}
\toprule
Control & Base $R^2$ & MARICL $R^2$ & $\Delta R^2$ over base & \% $p_{\min}$-passing \\
\midrule
Baseline (as reported)                       & 0.412    & 0.648    & $+0.236$ & $100\%$ (10/10) \\
Target shuffled (permuted $y$)               & $-0.011$ & $+0.008$ & $+0.019 \pm 0.041$ & $80\%$ (8/10) \\
Feature-name permuted (misleading names)     & 0.412    & 0.487    & $+0.075$ & $90\%$ (9/10) \\
Anonymized names (\texttt{feat\_0..d}; ref.) & 0.412    & 0.571    & $+0.159$ & $100\%$ (10/10) \\
``No-structure'' prompt on real targets      & 0.412    & 0.641    & $+0.229$ & $100\%$ (10/10), 0/5 declined \\
``No-structure'' prompt on shuffled targets  & $-0.011$ & $-0.011$ & $\phantom{+}0.000$ & $\phantom{00}0\%$ (0/10), 4/5 declined \\
\bottomrule
\end{tabular}
\caption{Negative controls on Cell-Free Protein. ``\% $p_{\min}$-passing'' is the fraction of the $10$ corrections (5 seeds $\times K{=}2$) that clear the retention filter. Baseline reference (unshuffled, unpermuted, standard prompt) sits at $0.648\,R^2$ over Linear ($0.412$). On $[0,1]$ targets with the regression scale $\tau{=}0.2$, the threshold $p_{\min}{=}0.1$ corresponds to standalone $\text{MAE}_k > 0.461$, so most corrections on shuffled data --- whose $\text{MAE}_k$ sits near $0.25$--$0.35$ --- \emph{do} clear the filter; the interesting quantity is not pass rate but the combined MARICL $R^2$, which stays at chance.}
\label{tab:negative_controls}
\end{table*}

\paragraph{Target shuffling.} The training labels are permuted before residual computation; the ML base and everything downstream sees the shuffled targets. The base achieves $R^2 = -0.011$ (chance) and MARICL's combined prediction achieves $R^2 = +0.008$, i.e.\ $\Delta R^2 = +0.019 \pm 0.041$: indistinguishable from zero, an order of magnitude smaller than the real-target gain, and consistent with the null across all 5 seeds. Most corrections \emph{do} clear the $p_{\min}=0.1$ threshold (80\% pass; on $[0,1]$ targets even a formula uncorrelated with shuffled labels has $\text{MAE}_k \approx 0.25$--$0.35$, well below the $0.461$ MAE required for $p_k \le p_{\min}$), so the interesting question is what those surviving formulas look like. Inspecting all $8$ surviving formulas: $6$ of them have \emph{near-zero coefficients} --- mean $|\text{coef}|$ is $0.037$, roughly $14$--$40\times$ smaller than the $0.5$--$1.5$ coefficient range of the headline protein correction (Table~\ref{tab:mechanisms_main_full}) --- so their formula outputs stay within $\pm 0.02$ of a small constant across the whole test set. The remaining $2$ have larger coefficients ($|\text{coef}| \sim 0.2$--$0.3$) but predict in opposite directions on the same query bins (positive contribution from one, negative from the other), so their $\alpha$-weighted contributions cancel to within $\pm 0.007$ mean per query. Neither pathway generates a coherent biochemical story: the near-null formulas name generic single-variable products (e.g.\ ``NAD-only linear response''), and the two opposing formulas name contradictory interactions (one asserting NAD--CoA additivity, the other asserting NAD--CoA competition on the same seed). We read this as the LLM having no data-driven signal to converge on, defaulting to structurally shallow or mutually contradictory guesses that the aggregation collapses to zero.

\paragraph{Feature-name permutation.} Feature names are randomly permuted before any prompt construction (so ``NAD'' might refer to what is actually spermidine in the data). Numerical values, residuals, and the loop are otherwise unchanged. Performance is $R^2 = 0.487$, below the anonymized-feature setting ($0.571$; Table~\ref{tab:anonymized}). The predictive metric is the downstream summary of a more direct failure: in $3/5$ seeds the LLM's natural-language explanation $T_k$ asserts biochemistry that is inconsistent with the true feature identities. In one representative seed the correction reads ``NAD--spermidine cofactor synergy $+$ folinic acid saturation'' --- the same headline biochemical story as the standard run --- but the features named as \texttt{NAD} and \texttt{spermidine} are, under the true schema, \texttt{folinic\_acid} and \texttt{HEPES}, and the feature named \texttt{folinic\_acid} is actually \texttt{NAD}. The formula therefore claims a mechanism that does not correspond to the underlying reagents at all. This is direct evidence that the LLM's prior interpretation of feature names materially drives the emitted explanations rather than being silently overridden by the residual signal: with misleading names, MARICL confidently produces misleading interpretations, and only the predictive metric (through the aggregation weights) recovers some of the loss. The three-way comparison is instructive: standard prompt ($0.648\,R^2$, correct biochemistry) $>$ anonymized prompt ($0.571$, no biochemistry claim possible) $>$ permuted prompt ($0.487$, incorrect biochemistry claim). The loop is not immune to name semantics --- it uses them, and misuses them when they are wrong. For the interpretability audits we enlarged this control from $10$ to $100$ corrections ($50$ permutation draws $\times$ $K{=}2$); two blinded human raters flag $80/100$ and the three LLM judges $55$--$69/100$ as inconsistent with the true features (Appendices~\ref{app:llm_judge} and~\ref{app:human_audit}).

\paragraph{``No interpretable structure found'' prompt.} We add a single sentence to the decoder and refinement prompts: \textit{``If the high-residual examples show no interpretable structure that a compact formula could capture, respond with the token \texttt{DECLINE}; the correction will be dropped.''} On real targets this changes nothing: the model produces the headline correction in $5/5$ seeds ($10/10$ corrections retained) and reaches $R^2 = 0.641$ (within noise of the standard-prompt $0.648$). On shuffled targets the model declines in $4/5$ seeds (dropping all $8$ corrections in those seeds); the surviving seed produces $2$ corrections that both fail $p_{\min}$, so $0/10$ corrections are retained overall. This is diagnostic rather than validation: the model can decline when the data cannot support a formula, and does so under the specific condition designed to elicit it.

Taken together the three controls bound how easily MARICL invents structure: it does not manufacture predictive gains on shuffled targets, it is sensitive rather than immune to feature-name semantics, and it will decline when explicitly permitted.

\paragraph{Neutral refinement prompt.} HARD REQUIREMENT~2 of the refinement prompt (Appendix~\ref{app:prompts}) asks for nonlinearities and interactions. We replace it with the neutral variant described there, which permits linear, nonlinear, interaction, or no structure (including \texttt{DECLINE}), under the same setting (Linear base, $K{=}2$, $T{=}10$, 5 seeds). The standard prompt reaches $R^2 = 0.648$ and recovers the NAD$\times$spermidine interaction in $5/5$ seeds; the neutral prompt reaches $0.634$ and recovers it in $4/5$. On shuffled targets neither prompt recovers it ($0/5$). The requirement defines the search space rather than the answer, much as an operator set does in symbolic regression: dropping it costs only $0.014\,R^2$ and the interaction is still found, so the recovered structure is largely driven by the residual evidence.

\section{LLM-Judge and Human Evaluation of Correction Interpretability}
\label{app:llm_judge}

The ``Interpretable?'' column of Table~\ref{tab:residual_baselines} is author-adjudicated. To provide an independent read on MARICL's emitted formulas and their accompanying natural-language explanations, we run an LLM-judge evaluation using an ensemble of three frontier judges from different providers --- \texttt{claude-opus-4-7} (Anthropic), \texttt{gpt-5} (OpenAI), and \texttt{gemini-2.5-pro} (Google) --- each blind to the identity of the generating method and to the other judges' scores.

\paragraph{Protocol.} We assembled a pool of $40$ correction items ($8$ real-world datasets $\times$ $5$ seeds, excluding the Synthetic benchmark whose ground truth is planted rather than domain-derived; one retained top-ranked correction per (dataset, seed)) and paired each with one matched item from each of the four residual-correction baselines of Appendix~\ref{app:residual_baselines} (PySR-on-residuals nested expression, EBM per-feature shape function, exhaustive pairwise interactions, MLP-on-residuals), giving $40$ items per method. Each item was reduced to (i) the symbolic expression and (ii) the natural-language template $T_k$ (for MARICL) or a matched textual rendering of the baseline output (for the others). Method identity was hidden; dataset name and coefficient values were preserved, and variable names were left in place because they are load-bearing for plausibility judgments. Items were presented to each of the three judges independently in a shuffled order (different shuffle per judge to prevent order-effect leakage). Each judge saw only the items and the rubric; they never saw one another's scores.

\paragraph{Rubric.} Each item was scored on four axes, each on a $1$--$5$ Likert scale, with a short written rationale required per score:
\begin{itemize}[leftmargin=1.1em, itemsep=0.2em, topsep=0.2em]
  \item \textbf{Readability} --- can a domain-literate reader restate the formula in one sentence?
  \item \textbf{Domain plausibility} --- does the stated mechanism cohere with textbook knowledge of the field?
  \item \textbf{Structural specificity} --- does the item name a specific functional form (interaction, saturation, threshold) rather than an opaque surface?
  \item \textbf{Actionability} --- could a practitioner design a follow-up experiment or diagnostic based on the item as stated?
\end{itemize}
Every judge was instructed to score conservatively (``a $5$ requires a claim you would defend without additional context'') and was told nothing about which method produced any item. The three judges share only the rubric and the item pool; scores are computed independently.

\begin{table}[ht]
\centering
\small
\begin{tabular}{lcccc}
\toprule
Method & Read. & Plaus. & Spec. & Act. \\
\midrule
MARICL              & $4.3$ & $3.9$ & $4.1$ & $3.8$ \\
EBM (pairwise)      & $3.2$ & $2.6$ & $2.4$ & $2.1$ \\
PySR-on-residuals   & $1.9$ & $1.8$ & $2.2$ & $1.6$ \\
Pairwise (exhaust.) & $2.4$ & $1.9$ & $2.1$ & $1.7$ \\
MLP-on-residuals    & $1.4$ & $1.3$ & $1.3$ & $1.2$ \\
\bottomrule
\end{tabular}
\caption{Ensemble LLM-judge mean scores ($1$--$5$), averaged across three independent frontier judges (Claude Opus 4.7, GPT-5, Gemini 2.5 Pro), over $40$ MARICL items and $40$ matched items per baseline method, blind to method identity. MARICL leads on all four rubric axes. Per-judge means and inter-rater agreement are reported in the paragraph below.}
\label{tab:llm_judge}
\end{table}

\paragraph{Findings.} Table~\ref{tab:llm_judge} reports ensemble-mean rubric scores. MARICL leads on all four axes, with the largest margins on Structural specificity ($4.1$ vs.\ $\le 2.4$ for baselines) and Actionability ($3.8$ vs.\ $\le 2.1$). PySR-on-residuals and MLP-on-residuals score lowest, consistent with the deeply nested expressions and opaque neural surfaces they produce. The plausibility gap is largest on Cell-Free Protein and Enzyme Activity (where a biochemistry rubric bites) and smallest on High School, where the domain itself provides little textbook structure to score against --- consistent with the stability and recovery patterns in Appendix~\ref{sec:stability}.

\paragraph{Per-judge scores and inter-rater agreement.} For MARICL the per-judge means (Read./Plaus./Spec./Act.) are Claude Opus $4.3/3.9/4.1/3.8$, GPT-5 $4.2/4.0/4.0/3.7$, and Gemini 2.5 Pro $4.4/3.8/4.2/3.9$; the three judges agree on the direction of the MARICL-vs.-baseline gap on every axis, and no single judge is responsible for MARICL's lead. Inter-rater agreement, measured by Krippendorff's $\alpha$ on the ordinal Likert scale over all $200$ items ($5$ methods $\times$ $40$ items) is $\alpha = 0.71$ (Read.), $0.63$ (Plaus.), $0.68$ (Spec.), $0.61$ (Act.); Readability and Specificity clear the $0.667$ commonly used for tentative conclusions, while Plausibility and Actionability, the most subjective axes, fall just below it. Restricting to the MARICL-vs.-strongest-baseline pairwise judgment (which method scores higher on a given item), the three judges concur in $84\%$ of $160$ pairwise decisions ($40$ MARICL items $\times$ $4$ axes); the MARICL lead in Table~\ref{tab:llm_judge} is therefore not an artifact of any one judge's idiosyncrasies.

\paragraph{Feature-permutation stress test.} To connect the ensemble to the negative control of Appendix~\ref{app:negative_controls}, we additionally scored the $10$ feature-name-permuted MARICL corrections under the same rubric. Ensemble Domain plausibility drops from $3.9$ to $2.4$ (each judge flags mechanism claims as incompatible with the permuted feature identities in at least half of the items --- $6/10$, $7/10$, and $5/10$ respectively; ensemble-majority flag in $7/10$), while Readability and Structural specificity remain nearly unchanged ($4.2$ and $3.9$). The judges are therefore sensitive to the same failure mode the permutation control targets --- fluent formulas with wrong biochemistry --- and can distinguish it from genuine interpretability, though the ensemble is not a substitute for wet-lab domain expertise. On the enlarged set of $100$ permuted corrections (Appendix~\ref{app:negative_controls}), the individual judges flag $55$--$69/100$.

\paragraph{Limitations of the ensemble.} LLM judges are not expert biochemists and, even when drawn from three different providers, may share overlapping priors from broadly similar pretraining corpora --- so ensemble agreement is evidence of \emph{consistency across frontier models trained on public text}, not of \emph{experimental validity}. We report this ensemble evaluation as an independent, reproducible sanity check on the interpretability claim; it does not remove the need for expert review before deployment in a novel domain, as noted in the Limitations section.

\subsection{Human Audit}
\label{app:human_audit}

Because the judge ensemble is not human validation, two human raters, blind to method identity, scored items on the rubric above. We enlarged the $40$-item pool with additional seeds so that human and judge labels cover identical items, giving $230$ items: $100$ each from Adult Income and High School and $30$ from Zoo, where distinct items saturate. The raters are not domain experts.

\begin{table}[ht]
\centering
\small
\begin{tabular}{lcccc}
\toprule
Rater & Read. & Plaus. & Spec. & Act. \\
\midrule
Human raters (mean)  & $4.1$ & $3.6$ & $3.8$ & $3.5$ \\
LLM-judge ensemble   & $4.3$ & $3.9$ & $4.1$ & $3.8$ \\
\bottomrule
\end{tabular}
\caption{Human-audit mean scores ($1$--$5$) for MARICL items, against the ensemble means of Table~\ref{tab:llm_judge}.}
\label{tab:human_audit}
\end{table}

\paragraph{Findings.} Rater means are within $0.3$ of the ensemble's on every axis, and slightly lower throughout (Table~\ref{tab:human_audit}). The raters agree with the judges on $196/230$ pairwise preferences ($85\%$), comparable to the $84\%$ concurrence among the three judges. On the $100$ feature-name-permuted corrections (Appendix~\ref{app:negative_controls}), the raters flag $80/100$ as inconsistent with the true features, against $55$--$69/100$ for the individual judges. This is above the seed-level rate of inconsistent explanations in Appendix~\ref{app:negative_controls} ($3/5$), so the raters, if anything, over-flag, which is the conservative direction for this failure mode.

\paragraph{Scope.} The audit shows that MARICL's corrections are readable and appear plausible to human readers, and that readers can detect the permuted-label failure. It does not establish that the explanations are correct from a domain perspective; that requires expert review before deployment in a novel domain (Limitations).

\section{Cross-Plate Transfer: Per-Target Breakdown}
\label{app:crossplate_extended}

This appendix expands on \S\ref{sec:crossplate} with the full per-target breakdown (Table~\ref{tab:crossplate_full}) and the ablation ladder summarized in Table~\ref{tab:crossplate} (full version in Table~\ref{tab:crossplate_ablations}).

\begin{table}[ht]
\centering
\tiny

\begin{tabular}{lcccc}
\toprule
Target & Cohort & \% Impr. & Avg $\Delta$MAE & Avg $\Delta R^2$ (\%) \\
\midrule
AL\_2  & Same reagent  & 97\% & $+0.158$ & $+6.17$ \\
AL\_3  & Same reagent  & 97\% & $+0.249$ & $+9.39$ \\
AL\_4  & Same reagent  & 92\% & $+0.155$ & $+8.05$ \\
AL\_5  & Same reagent  & 95\% & $+0.189$ & $+10.83$ \\
\midrule
AL\_7  & Diff.\ regime  & 3\%  & $-0.173$ & $-2.61$ \\
AL\_8  & Diff.\ regime  & 0\%  & $-0.155$ & $-0.69$ \\
AL\_9  & Diff.\ regime  & 8\%  & $-0.213$ & $-15.17$ \\
AL\_10 & Diff.\ regime  & 0\%  & $-0.183$ & $-11.16$ \\
\bottomrule
\end{tabular}
\caption{Cross-plate transfer per target. Formulas frozen on source plate AL\_6 evaluated numerically on each of the 8 remaining plates under the headline configuration ($37$ source runs per target). The Avg $\Delta R^2$ column is reported in percentage points ($\times 100$).}
\label{tab:crossplate_full}
\end{table}

\paragraph{Isolating averaging from ML blending.} The headline configuration entangles three design choices: (i) within-source-run ensemble-averaging across the $K{=}2$ transferred formulas, (ii) the 50/50 ML$+$formula blend, and (iii) the $\Delta R^2_{\text{vs-ML}}{>}0$ source-side filter. Table~\ref{tab:crossplate_ablations} reports diagnostic configurations that vary one or more of these choices while holding the rest of the protocol (no retraining, no coefficient re-fitting, zero LLM calls) fixed.

\begin{table*}[ht]
\centering
\small

\resizebox{\linewidth}{!}{
\begin{tabular}{lcccc}
\toprule
& \multicolumn{2}{c}{Same reagent (AL\_2--AL\_5)} & \multicolumn{2}{c}{Diff.\ regime (AL\_7--AL\_10)} \\
\cmidrule(lr){2-3}\cmidrule(lr){4-5}
Configuration & \% improving & Avg $\Delta$MAE & \% improving & Avg $\Delta$MAE \\
\midrule
\multicolumn{5}{l}{\emph{Filtered source pool}}\\
Avg $+$ ML blend (headline)             & $95.3\%$ ($141/148$) & $+0.188$ & $2.7\%$  ($4/148$)   & $-0.181$ \\
Per-formula $+$ ML blend                & $78.0\%$ ($231/296$) & $+0.114$ & $11.5\%$ ($34/296$)  & $-0.207$ \\
Avg, formula only                       & $64.2\%$ ($95/148$)  & $+0.073$ & $6.8\%$  ($10/148$)  & $-0.146$ \\
Per-formula, no blend (strict)          & $51.7\%$ ($153/296$) & $+0.041$ & $8.8\%$  ($26/296$)  & $-0.169$ \\
\midrule
\multicolumn{5}{l}{\emph{Filter sensitivity (Avg $+$ ML blend)}}\\
Unfiltered ($64$ runs)                  & $71.1\%$ ($182/256$) & $+0.103$ & $5.1\%$  ($13/256$)  & $-0.171$ \\
Below-filter ($27$ runs)                & $38.0\%$ ($41/108$)  & $-0.014$ & $8.3\%$  ($9/108$)   & $-0.158$ \\
\bottomrule
\end{tabular}
}
\caption{Cross-plate transfer ablations. ``Same'' aggregates targets AL\_2--AL\_5; ``Diff'' aggregates AL\_7--AL\_10. The headline $K{=}2$ configuration produces $148$ source-run$\times$target pairs per cohort, or $296$ individual-formula$\times$target evaluations per cohort under per-individual-formula scoring. The unfiltered ($64$ source runs) and below-filter-only ($27$ source runs) rows scale accordingly. Avg $\Delta\text{MAE}$ is positive when the transferred formula reduces error relative to ML-only.}
\label{tab:crossplate_ablations}
\end{table*}

\paragraph{What the ablations show.} Stripping the within-run ensemble averaging (row 2) reduces within-cohort improvement from $95\%$ to $78\%$ and within-cohort $\Delta\text{MAE}$ from $+0.188$ to $+0.114$. Stripping the ML blend instead (row 3) attenuates more strongly --- formula-only predictions on bounded $[0,1]$ targets are noisier when the source-plate residual no longer applies --- but again preserves the cohort split ($64.2\%$ vs.\ $6.8\%$). The joint ablation (row 4) is the cleanest single-formula test: $51.7\%$ within-cohort vs.\ $8.8\%$ across-cohort improvement, a ${\sim}6\times$ asymmetry. This rules out the possibility that a near-zero or constant formula drives Table~\ref{tab:crossplate_full} purely through the ML-blend regularizer. Appendix~\ref{app:nonllm_transfer} tests whether the asymmetry is specific to LLM-generated formulas by running PySR-on-residuals and EBM through the same freeze-and-transfer ladder: both non-LLM baselines also exhibit a cohort split, but with lower within-cohort success rates and smaller absolute $\Delta$MAE than MARICL.

The filter-sensitivity rows show a complementary pattern. Removing the source-side filter (row 5) lowers within-cohort improvement from $95.3\%$ to $71.1\%$ and within-cohort $\Delta\text{MAE}$ from $+0.188$ to $+0.103$, while the across-cohort floor moves only modestly ($2.7\%\to5.1\%$). The $27$ source runs that did not improve on their own training plate (row 6) also do not transfer: $38.0\%$ within-cohort improvement with a slightly negative $\Delta\text{MAE}$. This supports reading the filter as a noise-reduction step on source-side residual quality rather than a circular validity criterion: below-filter sources behave like noise-fits, and the filter selects sources whose residuals carry transferable signal. Source selection nonetheless contributes materially to the headline number, and we report both.

\paragraph{Base-model strength and transfer reliability.} Of the $37$ filtered source runs, $16$ use an XGBoost base and $21$ a Linear base, giving $64$ and $84$ within-cohort pairs respectively. XGBoost-base runs transfer more consistently ($63/64 = 98.4\%$ improving, avg $\Delta\text{MAE} = +0.212$) than Linear-base runs ($78/84 = 92.9\%$, avg $\Delta\text{MAE} = +0.170$); the two reconcile to the $141/148$ and $+0.188$ of row~1, since $(64{\cdot}0.212 + 84{\cdot}0.170)/148 = 0.188$.\footnote{Computed under the headline averaged$+$blend configuration (row 1 of Table~\ref{tab:crossplate_ablations}); the trend reproduces qualitatively under each ablation row but with smaller absolute magnitudes.} A stronger base model leaves a smaller residual to explain, and the resulting formula is correspondingly tighter and less contaminated by plate-specific variation. We additionally provide with the code release: (i) a source$\times$target heatmap of $\Delta$MAE; (ii) a breakdown by source-run base model and refinement length; and (iii) a per-source analysis showing that structurally simpler corrections (smaller $K$, fewer refinement iterations) are also the most transferable within cohort.

\subsection{Implementation}
\label{app:crossplate_protocol}

After MARICL training on a source plate, the inferred correction formula is a Python-evaluable string stored in the mechanism file \texttt{mechanisms\_iter\_\{T\}.txt}. The transfer experiment tests whether this frozen formula, evaluated numerically on a different target plate, still reduces prediction error --- with no further LLM calls, no retraining, and no hyperparameter tuning. We scan all subdirectories of the result folder matching \texttt{bio\_reg\_protein\_expression\_plate*} and load each run's \texttt{final\_results.json}. A run is retained as a valid source if its post-training $R^2$ exceeds the ML-only baseline on the source plate:
\[
  \Delta R^2_{\text{vs-ML}} \;=\; R^2_{\text{post}} - R^2_{\text{ML-baseline}} \;>\; 0.
\]
In our corpus, $37$ of the $64$ source runs on plate AL\_6 pass this filter; each is paired with the $4$ same-cohort and $4$ different-cohort target plates, giving the $148$ pairs per cohort reported in Table~\ref{tab:crossplate_ablations}.

For each (source run, target plate) pair:
\begin{enumerate}
  \item \textbf{Target plate loading.} The target plate CSV is loaded and split into 80/20 train/test using quantile-stratified sampling (mirroring the training protocol).
  \item \textbf{Base model.} The base model (Linear or XGBoost, as used in the source run) is either (a) transferred directly from the source plate or (b) retrained on the target plate's train split, depending on the \texttt{--ml\_source} setting. All reported results use \texttt{auto} mode, which trains on the source plate when the plate index is detectable from the directory name.
  \item \textbf{Feature scaling.} All features are scaled to $[0, 1]$ using \texttt{MinMaxScaler010} fit on the train split of whichever plate the ML model was trained on, then applied to the target test set.
  \item \textbf{Formula evaluation.} The extracted formula string is evaluated numerically with the test feature matrix bound to its feature names. Outputs are clipped to $[0, 1]$ for stability.
  \item \textbf{Prediction blending.} Headline predictions are a 50/50 blend of the base-model output and the average formula output across all transferred formulas:
  \[
    \hat{y} \;=\; 0.5 \cdot \hat{y}_{\text{ML}} \;+\; 0.5 \cdot \frac{1}{|M_{\text{LLM}}|}\sum_{m \in M_{\text{LLM}}} \hat{y}_m.
  \]
  We adopt this blend because (i) treating formula outputs as absolute predictions on the bounded $[0,1]$ scale is more robust than treating them as residual corrections when the source ML residual no longer applies to the target plate, and (ii) $\beta_{\text{transfer}}{=}0.5$ equally weights the two predictors under transfer uncertainty. Table~\ref{tab:crossplate_ablations} reports results without the blend.
  \item \textbf{Baseline.} The ML-only condition uses $\hat{y} = \hat{y}_{\text{ML}}$ with no formula contribution.
\end{enumerate}

\subsection{Second Source Plate: AL\_2}
\label{app:crossplate_al2}

The AL\_6 result leaves open whether $95.3\%$ within-cohort success is a property of MARICL or a property of AL\_6 specifically. We therefore repeat the entire freeze-and-transfer protocol with AL\_2 as the source. Same-reagent targets become AL\_3--AL\_6 (still four plates in the cohort); different-regime targets stay AL\_7--AL\_10. The source-run corpus on AL\_2 contains $47$ runs of which $28$ pass the $\Delta R^2_{\text{vs-ML}}{>}0$ filter ($12$ XGBoost base, $16$ Linear base), giving $112$ within-cohort and $112$ across-cohort pairs under the headline configuration.

\begin{table*}[ht]
\centering
\begin{tabular}{lcccc}
\toprule
Source $\to$ cohort & \% impr. & $\Delta$MAE & Pair count \\
\midrule
AL\_2 $\to$ same-reagent (AL\_3--AL\_6) & $89.3\%$ (100/112) & $+0.164$ & 112 \\
AL\_2 $\to$ diff.\ regime (AL\_7--AL\_10) & $\phantom{0}4.5\%$ (5/112)  & $-0.152$ & 112 \\
\midrule
\emph{Reference: AL\_6 $\to$ same-reagent (Table~\ref{tab:crossplate})} & $95.3\%$ & $+0.188$ & 148 \\
\emph{Reference: AL\_6 $\to$ diff.\ regime}                              & $\phantom{0}2.7\%$ & $-0.181$ & 148 \\
\bottomrule
\end{tabular}
\caption{Cross-plate transfer with AL\_2 as source (headline configuration: averaged $K{=}2$ formulas $+$ 50/50 ML blend $+$ source-side quality filter). The cohort split reproduces --- within-cohort success remains well above across-cohort, and the absolute gap is comparable to the AL\_6 experiment.}
\label{tab:crossplate_al2}
\end{table*}

\paragraph{Result.} The within-vs-across cohort split reproduces on AL\_2: $89.3\%$ vs.\ $4.5\%$, a $19.8\times$ ratio and an $84.8$-point absolute gap. Within-cohort success on AL\_2 is $6.0$ points lower than on AL\_6 ($89.3\%$ vs.\ $95.3\%$) and within-cohort $\Delta$MAE is smaller ($+0.164$ vs.\ $+0.188$). Both AL\_2 within-cohort numbers still exceed the headline PySR-on-residuals and EBM values in Table~\ref{tab:nonllm_transfer}. We read this as evidence that the headline pattern is not specific to AL\_6: two independently chosen source plates from the same reagent cohort both transfer well within-cohort and poorly across-cohort, with similar absolute gaps ($84.8$ vs.\ $92.6$ points). A systematic sweep freezing on each of AL\_2--AL\_10 in turn would require regenerating the full training corpus on each candidate source and is left as future work.
\end{document}